\pdfoutput=1
\documentclass[11pt]{article}
\usepackage[preprint]{acl}
\usepackage{stfloats} 
\usepackage{times}
\usepackage{latexsym}
\usepackage{amsmath}
\usepackage{paracol}
\usepackage{caption}
\usepackage{tabularx}
\usepackage[ruled,vlined,linesnumbered]{algorithm2e}
\usepackage{booktabs}
\usepackage{amssymb}
\usepackage{multirow}
\usepackage[most]{tcolorbox}
\tcbuselibrary{listings,breakable}
\usepackage{listings}
\usepackage{hyperref}
\usepackage[T1]{fontenc}
\usepackage[utf8]{inputenc}
\usepackage{microtype}
\usepackage{inconsolata}
\usepackage{graphicx}
\usepackage[caption=false,font=footnotesize]{subfig} 
\usepackage{url}
\usepackage{float} 
\usepackage{fancyhdr}
\usepackage{booktabs}
\usepackage{multirow}
\usepackage{courier}
\usepackage{pdflscape} 
\usepackage{booktabs}
\usepackage{multirow}
\usepackage{array}
\usepackage{makecell}

\title{Agent Bain vs. Agent McKinsey:\\
A New Text-to-SQL Benchmark for the Business Domain}

 \author{
Yue Li$^{\spadesuit}$\thanks{Equal contribution.}\thanks{Corresponding author.},
Ran Tao$^{\spadesuit}$\footnotemark[1],
Derek Hommel$^{\clubsuit}$,
Yusuf Denizay Dönder$^{\clubsuit}$,\\
\textbf{
Sungyong Chang$^{\spadesuit}$,
David Mimno$^{\spadesuit}$,
Unso Eun Seo Jo$^{\spadesuit}$$^{\clubsuit}$,
}\\
$^{\spadesuit}$Cornell University\\
$^{\clubsuit}$Gena AI\\[4pt]
\texttt{\{yl3865, rt529, sc3427, mimno, unsojo\}@cornell.edu}\\
\texttt{\{derek, denizay, unso\}@gena.com}
\footnotetext[1]{Equal contribution.}
}
\begin{document}
\maketitle

\begin{abstract}
Text-to-SQL benchmarks have traditionally only tested simple data access as a translation task of natural language to SQL queries. But in reality, users tend to ask diverse questions that require more complex responses including data-driven predictions or recommendations. Using the business domain as a motivating example, we introduce CORGI, a new benchmark that expands text-to-SQL to reflect practical database queries encountered by end users. CORGI is composed of synthetic databases inspired by enterprises such as DoorDash, Airbnb, and Lululemon. It provides questions across four increasingly complicated categories of business queries: descriptive, explanatory, predictive, and recommendational. This challenge calls for causal reasoning, temporal forecasting, and strategic recommendation, reflecting multi-level and multi-step agentic intelligence. We find that LLM performance degrades on higher-level questions as question complexity increases. CORGI also introduces and encourages the text-to-SQL community to consider new automatic methods for evaluating open-ended, qualitative responses in data access tasks. Our experiments show that LLMs exhibit an average 33.12\% lower success execution rate (SER) on CORGI compared to existing benchmarks such as BIRD, highlighting the substantially higher complexity of real-world business needs. We release the CORGI dataset, an evaluation framework, and a submission website to support future research.

\end{abstract}

\section{Introduction}
\begin{figure}[!t]
    \centering
    \includegraphics[width=0.95\columnwidth]{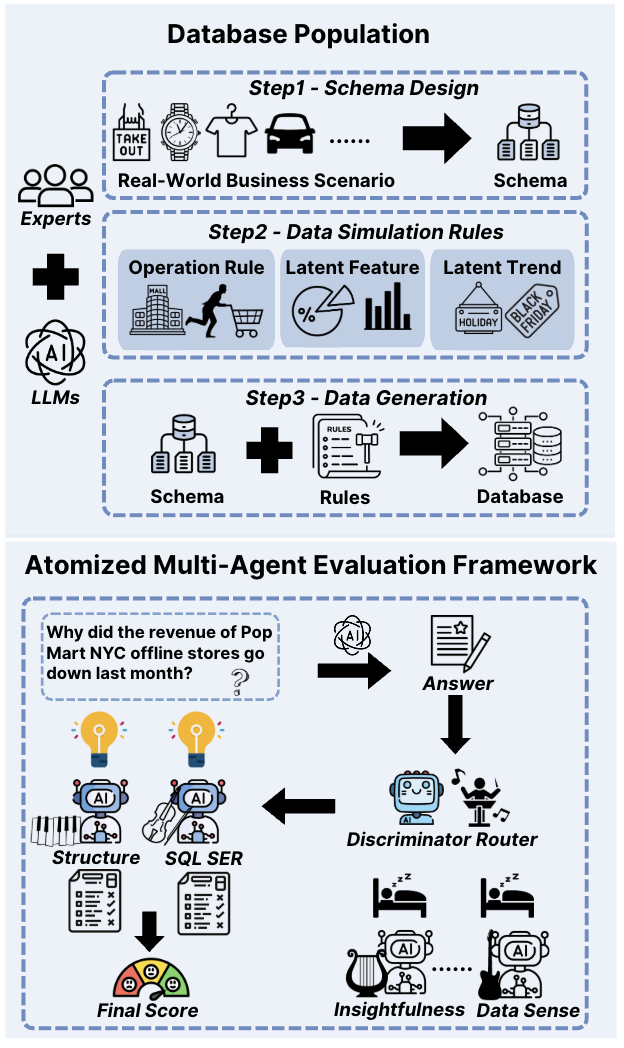}
    \caption{We incorporate business logic into the construction of CORGI and introduce a multi-agent evaluation framework with one discriminator and six scoring agents, each corresponding to a distinct scoring dimension for four types of business queries.}
    \label{fig:population-and-eval}
\end{figure}
Text-to-SQL is a long standing machine learning task of translating natural language questions (NLQ) into executable SQL queries \cite{zhong2017seq2seq,spider}. Most recently, LLMs have dominated state-of-the-art text-to-SQL approaches~\cite{sqlr1, nrep, arctictext2sqlr1} with commercial systems such as ChatGPT and Gemini making simple text-to-SQL functionality widely accessible. This calls into question the continued utility of text-to-SQL benchmarks, a task with a long tradition dating back to the 1970s, a much different era. Despite increasing complexity in joins and nested query structures, existing text-to-SQL benchmarks such as BIRD23-train-filtered~\cite{li2023bird}, WikiSQL~\cite{zhong2017seq2seq}, and Spider2.0~\cite{spider2} primarily focus on simple past data retrieval -- a translation task of mapping natural language input to SQL programming language. 

In practice, end users require more than querying past records; they also seek trend analysis, forecasting, and strategic recommendations, which calls for a broader conception of text-to-SQL that goes beyond pure translation and incorporates identifying relevant data and building data-driven responses~\cite{shmueli2010explain,fildes2008forecasting,armstrong2001principles,davenport2018ai}.
Focusing on business domain, we propose a more expansive paradigm for text-to-SQL benchmarking in the LLM era, emphasizing diverse user query types, LLM-generated qualitative responses, and automatic evaluation.
Our main contributions are as follows:

First, we present \textbf{CORGI}\footnote{In this paper, we use SQLite as the only SQL dialect.}, a business domain-specific text-to-SQL benchmark. CORGI provides simulated databases from verticals such as food delivery, rental, and advertising, as shown in Table~\ref{tab:company}, together with \textbf{four categories of business queries} including descriptive, explanatory, predictive, and recommendational. This captures real-world retrieval practices while enabling the evaluation of higher-level decision-making.

Second, we introduce a \textbf{business literature inspired evaluation framework} and an \textbf{atomized multi-agent evaluation mechanism}. 
Figure~\ref{fig:population-and-eval} illustrates the multi-agent evaluation mechanism. Based on the query type, the discriminator selects relevant scoring metrics and routes the input to specialized scoring agents, ensuring scalability and contextual precision. Table~\ref{tab:kendall_consistency} indicates that the scores given by LLMs under our framework are consistent with those of human experts.

Third, through preliminary experiments, we show that leading open-source and closed-source LLMs exhibit \textbf{performance drops} on explanatory, predictive or recommendational tasks, highlighting open challenges for future research. 

We release the CORGI dataset, code, evaluation questions, and evaluation framework on GitHub\footnote{\href{https://github.com/corgibenchmark/CORGI}{github.com/corgibenchmark/CORGI}}. We further provide a development set including gold labels for descriptive queries and human-annotated evaluation results for higher-level reasoning questions to facilitate community engagement.






\section{Related Work}
\subsection{Toward Business Intelligence (BI): Expanding the Scope of Text-to-SQL}
Earlier benchmarks such as WikiSQL~\cite{zhong2017seq2seq} and Spider~\cite{yu2018spider} predate the LLM era and primarily frame text-to-SQL as a translation task, often also called intent-to-SQL or NL-to-SQL. This has been primarily a simple task about accessing past data from databases. More recent benchmarks, including BIRD~\cite{li2023bird} and its extensions BIRD-CRITIC~\cite{li2025swe}, BIRD-Interact~\cite{huo2025bird}, and LiveSQLBench~\cite{livesqlbench2025} have been newly introduced, expanding evaluation to complex queries, SQL debugging, and agentic interaction, but still remain focused on text-to-SQL translation. 

However, real-world users require more than past data access. In business scenarios, database queries often demand forecasting and data-informed advice on topics such as pricing strategies and capacity planning~\cite{bughin2018notes, gathani2022whatif}. In healthcare, clinicians query patient records to predict risk and support treatment decisions~\cite{topol2019highperformance,shortliffe2021clinical}.

\subsection{Holistic Evaluation Framework}
Traditional text-to-SQL evaluation emphasizes the correctness of generated SQL queries and their execution results, typically measured by metrics such as SER and execution accuracy (EA)~\cite{wang-2019-glue,xu-2021-sql-eval}. However, as text-to-SQL is extended to BI settings, evaluation becomes increasingly subjective, since outputs often involve open-ended qualitative explanations rather than exact matches. Metrics like EA fail to capture key interpretability aspects, such as identifying and grounding relevant data, which are essential for downstream reasoning quality and decision relevance~\cite{bommasani-2021-foundation,zhou-2023-sql-limitations}. Prior work in business contexts further highlights that effective evaluation must go beyond technical correctness to account for strategic alignment, creativity, and judgment under uncertainty~\cite{iurii2024,jain2024,Malberg2025BridgingAI}.

Assessing such qualitative dimensions traditionally relies on human judgment, which is costly, inconsistent, and difficult to scale. While prior studies report mixed evidence on LLMs as evaluators~\cite{zheng2023mtbench}, other work shows that LLM-based judgments can closely align with human scoring when guided by explicit rubrics and applied to open-ended, reasoning-intensive tasks~\cite{liu2023geval}. Comparable performance has been observed in domains such as law, where LLMs evaluate complex reasoning tasks at near-expert levels~\cite{shui-2023-legal-ljp}. These advantages in scalability and consistency have driven the growing adoption of LLM-based evaluation, including in settings such as LLM alignment, where LLMs are used to assess the outputs of other models~\cite{anthropic2025bloom}.

\begin{figure*}[t]
    \centering
    \subfloat[Latent feature distribution (partial)\label{fig:latent-feature}]{%
        \includegraphics[height=5cm,keepaspectratio]{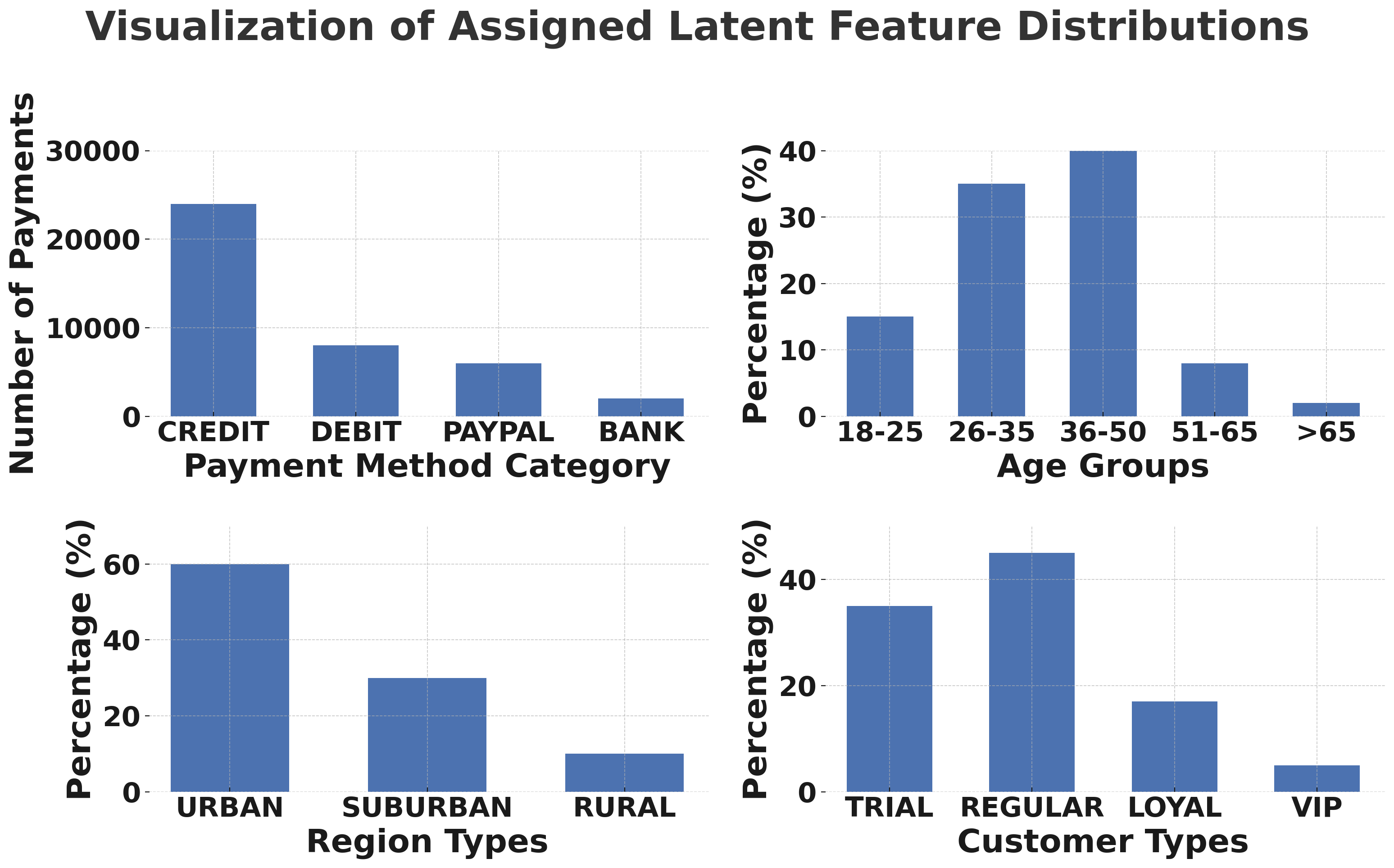}}
    \hfill
    \subfloat[Latent seasonal trends \label{fig:latent-trend}]{%
        \includegraphics[height=5cm,keepaspectratio]{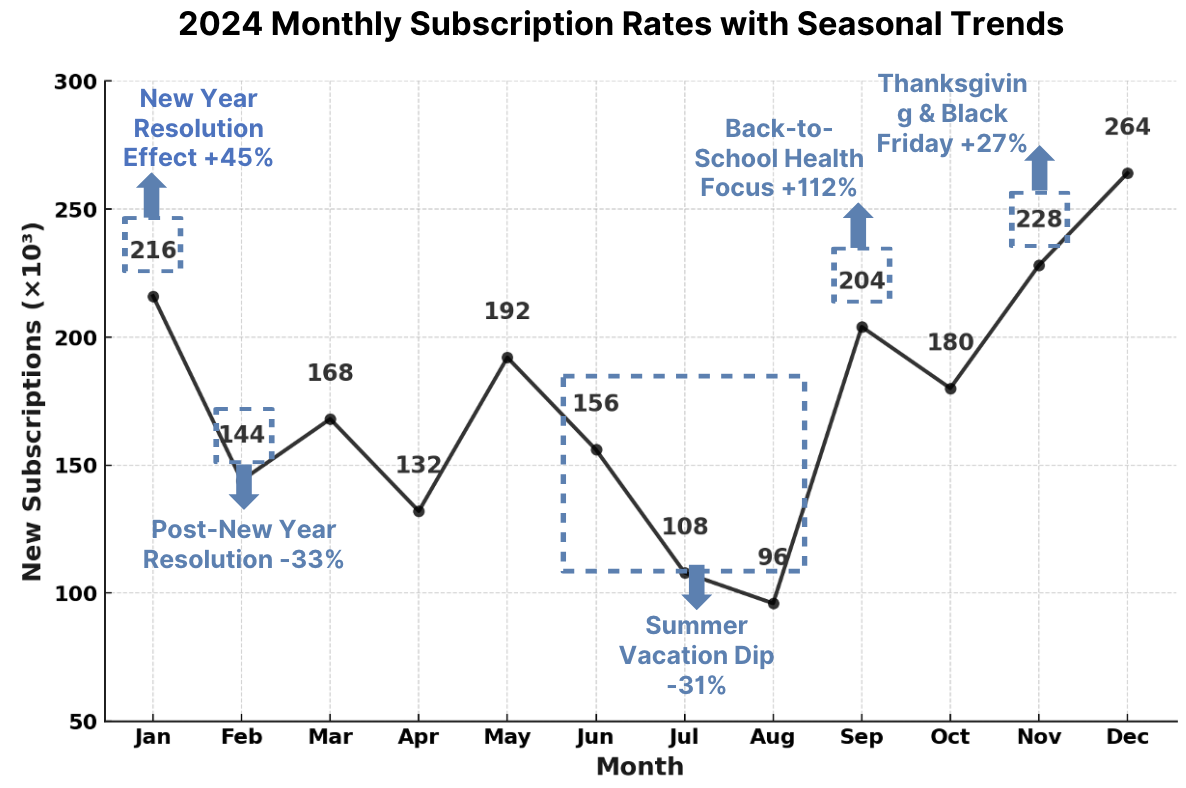}}
    \caption{Illustrations of data simulation rules without compounding effects, using the \textit{Persona Nutrition} database as a case study. Additional details are provided in Appendix~\ref{sec:data-generation-rules}.}
\end{figure*}

\section{Benchmark Construction}
\subsection{Database Population}
We first construct database schemas and then populate them with simulated data. We design the entity–relationship diagrams and schemas from publicly available sources, technical documentation, and application interfaces to emulate real-world business operations. 

We generate data using three types of rules~\footnote{We apply the three rule types independently as a simplifying assumption.}: (i) business operational constraints (e.g., the follow-up assessments conducted every six months), (ii) latent feature distributions (e.g., the customer type distribution illustrated in Figure~\ref{fig:latent-feature})
, and (iii) seasonal trends and fluctuations (e.g., the 45\% New Year resolution effect illustrated in Figure~\ref{fig:latent-trend}). Appendix~\ref{sec:data-generation-rules} provides detailed explanations and examples of these rules.

\begin{figure*}[t]
    \centering
    \includegraphics[width=\textwidth]{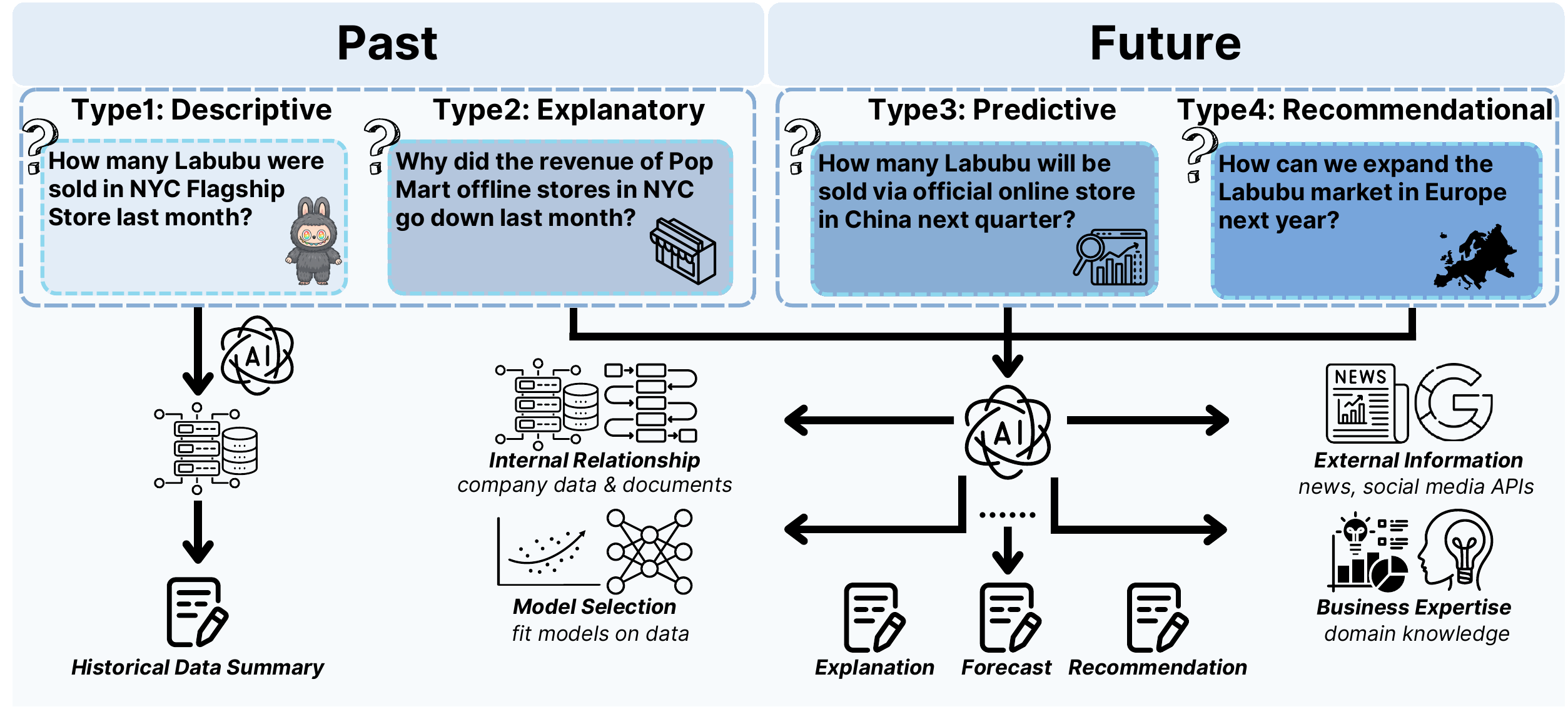}
    \caption{Queries range in complexity from fact retrieval to speculative business strategy.}
    \label{fig:4type}
\end{figure*}

\subsection{Question Generation}
Queries can be broadly categorized into two classes: those concerning the \textit{past} and those oriented toward the \textit{future}~\cite{shmueli2010explain,fildes2008forecasting}. 
Building on this distinction, we categorize queries into four types, as illustrated in Figure~\ref{fig:4type}. Overall, type~1 questions correspond to the classic text-to-SQL task, while the remaining types extend beyond pure translation and require \textit{data retrieval} to support higher-level responses.


\textbf{Type 1: Past, Descriptive.}
Type~1 questions focus on direct factual retrieval by translating NLQ into SQL~\cite{zhong2017seq2seq}.


\textbf{Type 2: Past, Explanatory.}
Type~2 queries require explaining historical patterns rather than merely retrieving data, necessitating appropriate SQL queries to identify relevant data as well as reasoning over underlying patterns~\cite{shmueli2010explain}.

Consider the question: ``Why did the revenue of Pop Mart’s offline stores in New York City decline over the past 90 days?'' Answering it requires analyzing \textit{retrieved sales and inventory data} to identify potential factors, such as supply chain disruptions.

\textbf{Type 3: Future, Predictive.}
Type~3 questions involve forecasting future outcomes from historical data~\cite{armstrong2001principles,fildes2008forecasting}, requiring trend identification and predictive model selection (e.g., time-series models~\cite{faw2025incontext}) based on SQL-retrieved data. 

For instance, the question ``What are the expected Labubu sales for the North America online store in November 2026?'' requires forecasting demand by combining \textit{historical sales trends} with upcoming promotions such as Black Friday.

\textbf{Type 4: Future, Recommendational.}
Type~4 queries require LLMs to generate actionable recommendations by identifying relevant historical records, evaluating alternative strategies, and proposing interventions aligned with business objectives~\cite{davenport2018ai,provost2013data}.

Consider the query: ``How can the Labubu market in Europe be expanded next year?'' A plausible answer can be derived by analyzing \textit{past offline revenue and conversion data} to suggest partnerships with high-performing retailers.

\subsection{CORGI Statistics}
CORGI covers 10 real-world industry verticals,\footnote{We consider only 10 industry domains in this paper, with additional ones left for future work.} spanning three broad categories: (i) consumer platforms (e.g., DoorDash), (ii) retail and commerce (e.g., Lululemon), and (iii) digital services (e.g., Apple App Store), as summarized in Table~\ref{tab:company}. Appendix~\ref{sec:appendix-business} provides the detailed descriptions.


Table~\ref{tab:query_stats} compares the overall and question-type-level structural complexity of CORGI with existing benchmarks, highlighting the substantially greater structural demands and answer-generation challenges posed by CORGI. Compared with BIRD, CORGI exhibits a markedly higher number of JOINs per question (7.48 vs.\ 0.93), more queries per question (2.31 vs.\ 1.0), and significantly longer answers (1002 vs.\ 161 characters). In addition, while the average SQL SER on the BIRD dev set reaches 88.0\%, the average SQL SER for type~2--4 questions in CORGI drops to 54.88\%, indicating increased difficulty in generating executable queries. Within CORGI itself, type~1 questions are markedly simpler than higher-level question types. Specifically, the generated SQL queries for type~2--4 questions are substantially longer on average (1159.33 vs.\ 255 characters) and involve more JOIN operations (3.80 vs.\ 0.55), further underscoring the challenges introduced by advanced BI-oriented tasks. Appendix~\ref{appen-sta} provides additional statistics of CORGI.


\begin{table*}[t]
\centering
\small
\begin{tabular}{l l l c c c}
\toprule
\textbf{Business Scenario} & \textbf{Representative Platform} & \textbf{Business Model} & \textbf{Tables} & \textbf{\#C/T} & \textbf{\#R} \\
\midrule
Food delivery              & DoorDash           & B2C/B2B2C  & 30  & 4.0   & 25 \\
Home rental                & Airbnb             & C2C  & 18  & 4.3   & 25 \\
Clothing e-commerce        & Lululemon          & B2C  & 34  & 3.9   & 40 \\
Freelancing marketplace    & Upwork             & C2C/C2B  & 25  &  3.8  & 33 \\
App store                  & Apple App Store        & B2C/B2B2C  & 18  & 4.3   & 22 \\
Car rental                 & Turo               & C2C  & 30  & 6.7  & 68 \\
Luxury consignment         & The RealReal       & C2C  & 30  & 6.5  & 65 \\
User analytics             & Amplitude          & B2B  & 20  & 7.5  & 25 \\
Personalized product       & Persona Nutrition  & B2C  & 28  & 7.8  & 38 \\
E-commerce enablement      & Shopify            & B2B  & 23  & 7.0  & 60 \\
\bottomrule
\end{tabular}
\caption{Overview of the ten industry verticals in CORGI. \#C/T and \#R denote the average number of columns per table and the number of foreign key relationships.}
\label{tab:company}
\end{table*}

\begin{table*}[!t]
\centering
\small
\begin{tabular}{lcccccccc}
\toprule
\textbf{Type} & \textbf{\#Q} & \textbf{J/Q} & \textbf{Q/Qs} & \textbf{J/Qr} & \textbf{C/Qr} & \textbf{G/Qr} & \textbf{S/Qr} & \textbf{Len} \\
\midrule
\multicolumn{9}{l}{\textbf{CORGI}} \\
CORGI-Type1 & 300 & 0.55  & 1.00 & 0.55 & 1.78 & 0.16 & 0.05 & 255  \\
CORGI-Type2 & 300 & 11.07 & 2.83 & 3.99 & 1.66 & 1.54 & 0.71 & 1152 \\
CORGI-Type3 & 300 & 8.96  & 3.37 & 2.65 & 1.52 & 1.78 & 0.28 & 1036 \\
CORGI-Type4 & 300 & 16.26 & 3.34 & 4.76 & 1.07 & 2.25 & 0.50 & 1290 \\
\midrule
\textbf{CORGI-Overall} & 1200 & 7.48 & 2.31 & 3.24 & 1.47 & 1.58 & 0.41 & 1002 \\
\midrule
\multicolumn{9}{l}{\textbf{Other Benchmarks}} \\
BIRD-Simple        & -- & 0.75 & 1.00 & 0.75 & 1.27 & 0.05 & 0.00 & 129  \\
BIRD-Moderate      & -- & 1.16 & 1.00 & 1.16 & 1.73 & 0.12 & 0.01 & 193  \\
BIRD-Challenging   & -- & 1.35 & 1.00 & 1.35 & 1.75 & 0.13 & 0.07 & 271  \\
BIRD-Overall\footnotemark[1] & 12751 & 0.93 & 1.00 & 0.93 & 1.46 & 0.08 & 0.01 & 161  \\
Spider\footnotemark[2] & 1034 & 0.50 & 1.00 & 0.50 & 0.57 & 0.27 & 0.00 & 107  \\
WikiSQL\footnotemark[3] & 80654 & 0.00 & 1.00 & 0.00 & 1.38 & 0.00 & 0.00 & 70   \\
Advising\footnotemark[4] & 4570 & 0.79 & 1.00 & 0.79 & 4.15 & 0.06 & 0.00 & 488  \\
LiveSQLBench-Base-Lite\footnotemark[5] & 270 & 1.17 & 1.00 & 1.17 & 0.43 & 0.57 & 0.10 & 527 \\
\bottomrule
\end{tabular}
\caption{
Structural complexity comparison between CORGI and the development sets of existing benchmarks. 
\#Q denotes the number of questions; J/Q denotes JOINs per question; Q/Qs denotes queries per question; J/Qr denotes JOINs per query; 
C/Qr denotes conditions per query; G/Qr denotes GROUP BY clauses per query; S/Qr denotes subqueries per query; and Len denotes query length in characters.}
\label{tab:query_stats}
\end{table*}

\footnotetext[1]{BIRD~\cite{li2023bird}}
\footnotetext[2]{Spider~\cite{spider}}
\footnotetext[3]{WikiSQL~\cite{zhong2017seq2seq}}
\footnotetext[4]{Advising~\cite{advising}}
\footnotetext[5]{LiveSQLBench-Base-Lite~\cite{livesqlbench2025}. We select BI questions in LiveSQLBench-Base-Lite and generate answers using GPT-5.}


\section{Evaluation Framework}
\label{evaluation}
\subsection{Motivation}

We introduce the concept of automatic qualitative evaluation for text-to-SQL tasks, using LLMs as evaluators. Although the traditional formulation of text-to-SQL often assumes that gold labels are unambiguous, even seemingly simple fact-retrieval queries can admit multiple valid interpretations. For example, a query such as ``how many students do I have?'' may refer to (i) the number of students currently enrolled in a school, (ii) the total number of students previously taught, or (iii) the number of students in a specific class. Similarly, the query ``what were my sales this month?'' could correspond to (i) total revenue, (ii) total items sold, (iii) total profit (revenue minus cost), or (iv) profit after tax. Prior text-to-SQL benchmarks, such as BIRD, explicitly provide additional context or constraints to resolve such ambiguities and enforce a single intended interpretation~\cite{li2023bird}. In contrast, our expanded text-to-SQL benchmark produces qualitative, open-ended responses that cannot be adequately evaluated using gold labels alone. To address this challenge, we propose evaluating BI-related qualitative text-to-SQL outputs with LLMs that approximate human judgment, drawing inspiration from how responses are assessed in real-world management consulting settings.

\subsection{Strategic Management Consulting--Inspired Evaluation Framework} 
SQL-based data access is central to business management consulting~\cite{brynjolfsson2011strengthening,bughin2018notes}. 
To evaluate LLMs in text-to-SQL-centric business decision-making contexts, we integrate this criteria for assessing professional strategic consulting with recent work on LLM-based decision-making agents~\cite{greiner2005consulting,christensen2013consulting,park2023generative,yao2023tree}. This framework is collaboratively developed with business school faculty, professional consultants, and corporate practitioners.

We organize the framework into \textbf{six} main dimensions:(1) Structure \cite{Minto2009}, (2) Data Sense \cite{ChambersMullickSmith1971, BinderWatkins2024}, (3) Insightfulness \cite{BinderWatkins2024, BacklerEtAl2025}, (4) Operational Implementability \cite{ChambersMullickSmith1971, Garner2013}, (5) Purpose Alignment \cite{PorterKramer2011}, and (6) Compliance \cite{CowgillEtAl2020, guo2025flamefinanciallargelanguagemodel}. Appendix~\ref{consulting-rubic} elaborates each dimension via sub-metrics. 

Among the six dimensions, some are universal, while others are question-type specific. For example, \textit{Structure} applies to all questions, requiring clear organization and coherent argumentation. In contrast, \textit{Compliance} is specific to type~4 questions, where risk management, regulatory, and ethical considerations are essential for recommendations.

We apply the proposed evaluation framework to higher-level type~2--4 questions. We use SQL SER to evaluate performance across all question types, and additionally report EA for type~1 questions. Table~\ref{tab:ser-ea} reports the values of SER and EA.



\subsection{Atomized Multi-Agent Evaluation Mechanism}
Consistent with the evaluation framework, our multi-agent evaluation mechanism consists of one discriminator and six scoring agents, as illustrated in Figure~\ref{fig:population-and-eval}. The discriminator look-up serves as a routing component for evaluation. 
By allowing agents to specialize in distinct evaluation dimensions, this architecture mirrors human committee-based assessment, where diverse perspectives improve reliability and validity~\cite{chan2023chatevalbetterllmbasedevaluators,rasheed2024largelanguagemodelevaluation}.

The evaluation proceeds as follows:

\textbf{(1) Input}: Each input consists of the question type, question, and corresponding answer.
\textbf{(2) Discriminator Routing}: The discriminator identifies relevant scoring dimensions based on the question type and content.
\textbf{(3) Scoring}: General and specialized scoring agents assign scores along relevant dimensions.
\textbf{(4) Aggregation}: The final score is the average of the dimension scores.

Table~\ref{tab:big-eval-3evals} indicates that LLMs do not exhibit a clear self-preference bias, i.e., a tendency to assign systematically higher scores to outputs generated by the same model. For instance, on type~2 questions, GPT-5 assigns an average score of 3.87 to its own outputs, while Gemini-2.5-Pro and Llama~4 assign 4.16 when evaluating GPT-5's responses.



\subsection{Automatic and Human Evaluation Consistency}
In addition to automatic LLM-based evaluation, we conduct a human expert assessment on a randomly selected subset of the benchmark by sampling 3 of the 10 databases (AppStore, Clothing, and Shopify) and manually scoring the corresponding LLM-generated answers using the same evaluation framework.

Table~\ref{tab:kendall_consistency} reports the consistency between automatic evaluation and human annotations using Kendall’s $\tau$\footnote{An introduction to Kendall’s $\tau$ is provided in Appendix~\ref{appendix:kendall}.}. Overall, the agreement is moderate and decreases with question complexity, from 0.52--0.71 (type~2) to 0.43--0.64 (type~3) and 0.31--0.57 (type~4), reflecting the increased difficulty of evaluating qualitative, judgment-intensive responses.

\begin{table*}[!t]
\centering
\small
\begin{tabular}{llccc}
\toprule
\textbf{Type} & \textbf{Generator\textbackslash   Evaluator} & \textbf{Gemini-2.5-Pro} & \textbf{GPT5} & \textbf{Llama4} \\
\midrule
\multirow{4}{*}{Type 2}
 & Gemini-2.5-flash-lite & 0.62 & 0.66 & 0.57 \\
 & Gemini-2.5-Pro & 0.67 & 0.68 & 0.58 \\
 & GPT5           & 0.71 & 0.69 & 0.62 \\
 & Llama4         & 0.53 & 0.61 & 0.63 \\
 & Qwen3          & 0.57 & 0.65 & 0.52 \\
\midrule
\multirow{4}{*}{Type 3}
 & Gemini-2.5-flash-lite & 0.58 & 0.52 & 0.48 \\
 & Gemini-2.5-Pro & 0.63 & 0.59 & 0.54 \\
 & GPT5           & 0.51 & 0.64 & 0.46 \\
 & Llama4         & 0.54 & 0.47 & 0.55 \\
 & Qwen3          & 0.56 & 0.43 & 0.41 \\
\midrule
\multirow{4}{*}{Type 4}
 & Gemini-2.5-flash-lite & 0.45 & 0.43 & 0.35 \\
 & Gemini-2.5-Pro & 0.41 & 0.43 & 0.37 \\
 & GPT5           & 0.46 & 0.41 & 0.31 \\
 & Llama4         & 0.57 & 0.38 & 0.43 \\
 & Qwen3          & 0.38 & 0.36 & 0.32 \\
\bottomrule
\end{tabular}
\caption{Kendall's tau consistency scores between generators (rows) and evaluators (columns) on CORGI.}
\label{tab:kendall_consistency}
\end{table*}

\section{Methodology}
\subsection{Problem Formulation}
Given an NLQ $Q$, a database schema $D$, and a knowledge set $K=\{K_I, K_E\}$, where $K_I$ denotes internal knowledge (e.g., database descriptions) and $K_E$ denotes external knowledge (e.g., information retrieved via web search APIs), an LLM parameterized by $\theta$ consumes prompts $P$ to generate a set of SQL queries $S$, their execution results $R$, and a final answer $A$. Algorithm~1 summarizes the procedure, and we illustrate the process below.

\textbf{Query generation.}
$h(\cdot\mid\theta)$ maps $(Q,D,K,P)$ to a list of SQL queries $S$. For type~1 questions, the query generator produces a single SQL query. For the others, it generates up to five queries to support complex reasoning:
\[
S = h(Q,D,K,P\mid\theta).
\]

\textbf{Query execution.}
$e(\cdot)$ runs the generated queries on the database defined by $D$ to obtain execution results $R$:
\[
R = e(S).
\]

\textbf{Answer composition.}
$f(\cdot\mid\theta)$ produces the final answer from the question $Q$, execution results $R$, and knowledge set $K$:
\[
A = f(Q,R,K\mid\theta).
\]

\textbf{Output.}
The system returns the answer $A$ with index-aligned $(S,R)$ pairs, discarding erroneous queries or replacing them with empty results.

\begin{algorithm}[t]
\caption{Query Generation and Answer Synthesis for a Single NLQ}
\label{alg:noniter}
\DontPrintSemicolon

\textbf{Initialize} $S \gets [\;]$, $R \gets [\;]$\;
$S \gets h(Q, D, K, P \mid \theta)$\;

\For{$s \in S$}{
    $r \gets e(s)$\;
    \If{$r$ is invalid}{
        $r \gets \emptyset$\;
    }
    Append $r$ to $R$\;
}

$A \gets f(Q, R, K \mid \theta)$\;
\Return $A, S, R$\;
\end{algorithm}

\subsection{Scoring Strategy}
We adopts the following two-stage scoring strategy:

\textbf{Step 1: Dimension scoring.}
Each answer is evaluated along the relevant dimensions, using a discrete 0–3–5 scale where 5 denotes excellent performance meeting professional consulting standards and 0 indicates critical failure.

\textbf{Step 2: Score aggregation.}
The final score is the average over all applicable dimensions:
\[
\text{Final Score} = \frac{1}{n}\sum_{i=1}^{n} \text{Dimension Score}_i,
\]
where $n$ is the number of relevant dimensions. Alternative aggregation strategies are discussed in Appendix~\ref{aggregation}.

\section{Experiments}
\subsection{Experimental Setup}
For the benchmark experiments, we evaluate a mix of popular open-source and closed-source LLMs. 
For answer generation, we use \textit{Gemini-2.5-flash-lite}~\cite{comanici2025gemini25pushingfrontier}, 
\textit{Gemini-2.5-Pro}~\cite{Gemini2.5Pro_2025}, 
\textit{GPT-5}~\cite{openai_gpt5}, 
\textit{Llama~4} (model ID: \texttt{meta-llama/Llama-4-Scout-17B-16E-\\
Instruct})~\cite{meta_llama4}, and \textit{Qwen~3} (model ID: \texttt{Qwen/Qwen3-235B-A22B-Instruct-\\
2507})~\cite{yang2025qwen3}.
For automatic evaluation, we use \textit{Gemini-2.5-Pro}, \textit{GPT-5}, and \textit{Llama~4}. We set the \textit{temperature} to 0 for all models to reduce sampling variance and improve reproducibility~\cite{BIRD-dataset}. For LLM-based automatic evaluation, we batch inputs by question type. Appendix~\ref{sec:appendix-prompts} presents prompts used for answer generation and evaluation. 

We randomly sample 3 out of the 10 databases (AppStore, Clothing, and Shopify). We provide gold labels for type~1 queries and conduct LLM-based automatic evaluation as well as human expert annotation on type~2--4 questions using the framework introduced in Section~\ref{evaluation}.



\begin{table*}[t]
\centering
\small
\begin{tabular}{lccccccc}
\toprule
\textbf{Type} & \textbf{Metric} & \textbf{G5} & \textbf{G2.5P} & \textbf{G2.5F} & \textbf{L4} & \textbf{Q3} & \textbf{Overall} \\
\midrule
Type~1 & EA & 88.89\% & 81.48\% & 80.62\% & 66.67\% & 29.63\% & 69.46\% \\
Type~1 & SER & 92.59\% & 88.89\% & 92.59\% & 85.19\% & 40.74\% & 79.47\% \\
Type~2 & SER & 83.33\% & 71.11\% & 65.30\% & 27.66\% & 48.39\% & 59.16\% \\
Type~3 & SER & 63.79\% & 53.49\% & 67.91\% & 56.52\% & 48.98\% & 58.14\% \\
Type~4 & SER & 58.62\% & 50.00\% & 70.18\% & 34.78\% & 23.08\% & 47.33\% \\
\bottomrule
\end{tabular}
\caption{EA is reported for type~1 questions, while SER is reported across all question types and models. G5 denotes GPT-5; G2.5P denotes Gemini-2.5-Pro; G2.5F denotes Gemini-2.5-Flash-Lite; L4 denotes Llama~4; and Q3 denotes Qwen~3. \textit{Overall} reports the average values across all models.}
\label{tab:ser-ea}
\end{table*}

\begin{table*}[t]
\centering
\vspace{-2pt}
\resizebox{\linewidth}{!}{
\begin{tabular}{ll*{7}{ccc}c}
\toprule
\multirow{2}{*}{\textbf{Type}} &
\multirow{2}{*}{
\makecell[l]{\textbf{Generator} \textbf{\textbackslash\ Evaluator}}
}
&
\multicolumn{3}{c}{\textbf{S}} &
\multicolumn{3}{c}{\textbf{DS}} &
\multicolumn{3}{c}{\textbf{I}} &
\multicolumn{3}{c}{\textbf{OI}} &
\multicolumn{3}{c}{\textbf{PA}} &
\multicolumn{3}{c}{\textbf{C}} &
\multicolumn{3}{c}{\textbf{Avg.}} &
\multirow{2}{*}{\textbf{Overall}} \\
\cmidrule(lr){3-5}\cmidrule(lr){6-8}%
\cmidrule(lr){9-11}\cmidrule(lr){12-14}%
\cmidrule(lr){15-17}\cmidrule(lr){18-20}%
\cmidrule(lr){21-23}
& &
\textbf{G2.5P} & \textbf{G5} & \textbf{L4}
& \textbf{G2.5P} & \textbf{G5} & \textbf{L4}
& \textbf{G2.5P} & \textbf{G5} & \textbf{L4}
& \textbf{G2.5P} & \textbf{G5} & \textbf{L4}
& \textbf{G2.5P} & \textbf{G5} & \textbf{L4}
& \textbf{G2.5P} & \textbf{G5} & \textbf{L4}
& \textbf{G2.5P} & \textbf{G5} & \textbf{L4}
& \\
\midrule

\multicolumn{24}{l}{\textbf{Type 2}} \\

& Gemini-2.5-flash-lite
& 3.85 & 3.80 & 3.55
& 4.05 & 3.97 & 4.10
& 2.95 & 3.39 & 3.45
& ---  & ---  & ---
& ---  & ---  & ---
& ---  & ---  & ---
& 3.65 & 2.95 & 3.70
& 3.43 \\

& Gemini-2.5-Pro
& 4.02 & 3.31 & 3.82
& 4.23 & 3.23 & 4.77
& 2.97 & 3.17 & 4.20
& ---  & ---  & ---
& ---  & ---  & ---
& ---  & ---  & ---
& 3.74 & 3.24 & 4.26
& 3.75 \\

& GPT-5
& 4.26 & 3.36 & 3.75
& 4.80 & 4.33 & 4.60
& 3.43 & 3.90 & 4.13
& ---  & ---  & ---
& ---  & ---  & ---
& ---  & ---  & ---
& 4.17 & 3.86 & 4.16
& 4.06 \\

& Llama~4
& 1.71 & 1.53 & 3.58
& 2.20 & 1.57 & 3.77
& 1.63 & 1.30 & 2.80
& ---  & ---  & ---
& ---  & ---  & ---
& ---  & ---  & ---
& 1.85 & 1.47 & 3.38
& 2.23 \\

& Qwen~3
& 1.96 & 1.60 & 2.44
& 1.50 & 0.87 & 1.87
& 0.67 & 0.60 & 1.20
& ---  & ---  & ---
& ---  & ---  & ---
& ---  & ---  & ---
& 1.37 & 1.02 & 1.84
& 1.41 \\




\midrule
& \textit{Closed-source Overall}
& & 3.75 &
& & 4.23 &
& & 3.51 &
& & --- &
& & --- &
& & --- &
& & 3.75 &
& \textbf{3.75} \\

& \textit{Open-source Overall}
& & 2.14 &
& & 1.96 &
& & 1.37 &
& & --- &
& & --- &
& & --- &
& & 1.82 &
& \textbf{1.82} \\

\midrule
\multicolumn{24}{l}{\textbf{Type 3}} \\

& Gemini-2.5-flash-lite
& 3.55 & 3.10 & 3.60
& 3.95 & 3.10 & 4.10
& 2.65 & 2.30 & 3.10
& ---  & ---  & ---
& ---  & ---  & ---
& ---  & ---  & ---
& 3.38 & 2.83 & 3.60
& 3.27 \\

& Gemini-2.5-Pro
& 4.31 & 3.69 & 4.04
& 4.89 & 3.47 & 4.57
& 3.07 & 2.83 & 4.17
& ---  & ---  & ---
& ---  & ---  & ---
& ---  & ---  & ---
& 4.09 & 3.33 & 4.26
& 3.89 \\

& GPT-5
& 4.11 & 3.80 & 4.02
& 5.00 & 4.64 & 4.78
& 3.37 & 3.67 & 3.93
& ---  & ---  & ---
& ---  & ---  & ---
& ---  & ---  & ---
& 4.16 & 4.04 & 4.24
& 4.15 \\

& Llama~4
& 2.58 & 2.31 & 3.95
& 3.23 & 2.24 & 4.16
& 2.13 & 1.87 & 3.50
& ---  & ---  & ---
& ---  & ---  & ---
& ---  & ---  & ---
& 2.65 & 2.14 & 3.87
& 2.89 \\

& Qwen~3
& 2.29 & 2.11 & 3.09
& 1.84 & 1.69 & 2.62
& 0.93 & 1.33 & 2.13
& ---  & ---  & ---
& ---  & ---  & ---
& ---  & ---  & ---
& 1.69 & 1.71 & 2.61
& 2.00 \\

\midrule

& \textit{Closed-source Overall}
&& 3.80&
&& 4.28&
&& 3.23&
&& ---&
&& ---&
&& ---&
&& 3.77&
& \textbf{3.77} \\

& \textit{Open-source Overall}
&& 2.72&
&& 2.63&
&& 1.98&
&& ---&
&& ---&
&& ---&
&& 2.45&
& \textbf{2.45} \\



\midrule

\multicolumn{24}{l}{\textbf{Type 4}} \\

& Gemini-2.5-flash-lite
& 3.20 & 2.75 & 3.30
& 4.10 & 3.20 & 4.30
& 2.70 & 2.30 & 3.10
& 3.30 & 2.80 & 3.50
& 3.55 & 2.90 & 3.70
& 1.90 & 1.60 & 2.10
& 3.13 & 2.76 & 3.33
& 3.12 \\

& Gemini-2.5-Pro
& 3.95 & 3.04 & 3.77
& 5.00 & 3.49 & 5.00
& 3.60 & 3.00 & 4.87
& 4.03 & 3.60 & 4.53
& 4.60 & 3.27 & 4.70
& 3.27 & 2.44 & 4.73
& 4.08 & 3.14 & 4.60
& 3.94 \\

& GPT-5
& 4.33 & 3.49 & 3.44
& 5.00 & 4.80 & 5.00
& 3.53 & 3.00 & 4.27
& 4.57 & 4.33 & 4.93
& 4.67 & 3.70 & 4.67
& 3.38 & 2.73 & 4.69
& 4.25 & 3.68 & 4.50
& 4.14 \\

& Llama~4
& 2.76 & 2.65 & 4.29
& 4.17 & 2.97 & 4.93
& 0.01 & 1.20 & 3.00
& 2.07 & 1.80 & 2.97
& 4.20 & 3.10 & 4.73
& 1.40 & 1.33 & 3.44
& 2.43 & 2.17 & 3.89
& 2.83 \\

& Qwen~3
& 1.69 & 1.78 & 2.35
& 2.49 & 2.10 & 2.67
& 0.01 & 1.33 & 1.60
& 1.37 & 1.03 & 2.27
& 2.33 & 1.70 & 2.60
& 1.82 & 1.42 & 2.91
& 1.62 & 1.56 & 2.40
& 1.86 \\


\midrule
& \textit{Closed-source Overall}
& & 3.47 & 
& & 4.43 & 
& & 3.37 & 
& & 3.95 & 
& & 3.97 & 
& & 2.98 & 
& & 3.72 &
& \textbf{3.72} \\


& \textit{Open-source Overall}
& & 2.59 & 
& & 3.22 & 
& & 1.19 & 
& & 1.92 & 
& & 3.11 & 
& & 2.05 & 
& & 2.35 &
& \textbf{2.35} \\

\bottomrule
\end{tabular}}
\caption{
Evaluation scores for type~2--4 questions under the proposed framework. S denotes Structure; DS denotes Data Sense; I denotes Insightfulness; OI denotes Operational Implementability; PA denotes Purpose Alignment; C denotes Compliance; G2.5P denotes Gemini-2.5-Pro; G5 denotes GPT-5; and L4 denotes Llama~4. Dashes indicate metrics not applicable to a given question type. \textit{Overall} reports the average values across all models.
}
\label{tab:big-eval-3evals}
\end{table*}


\subsection{Experimental Results and Discussion}
For type~1 questions, we evaluate models using SQL SER and EA. For type~2--4 questions, we evaluate models using SQL SER together with the strategic management consulting–inspired evaluation framework. Tables~\ref{tab:ser-ea} and~\ref{tab:big-eval-3evals} present the experimental results.




\paragraph{Finding 1: Closed-source LLMs perform well on type~1 questions.}
Closed-source models demonstrate consistently stronger performance on type~1 questions, achieving EA scores of 88.89\% (GPT-5), 81.48\% (Gemini-2.5-Pro), and 80.62\% (Gemini-2.5-Flash-Lite), alongside SER scores of 92.59\% (GPT-5), 88.89\% (Gemini-2.5-Pro), and 92.59\% (Gemini-2.5-Flash-Lite). In contrast, open-source models exhibit noticeably lower performance, with EA scores of 66.67\% (Llama~4) and 29.63\% (Qwen~3), and SER scores of 85.19\% (Llama~4) and 40.74\% (Qwen~3). Overall, these results suggest that state-of-the-art closed-source LLMs are better equipped to generate correct and executable SQL queries for simple factual retrieval tasks.

\paragraph{Finding 2: LLMs exhibit a monotonic decrease in average SER from type~1 to type~4 questions.}
Table~\ref{tab:ser-ea} reports SER values across all question types and models. For type~2--4 questions, we allow up to five generated queries to support complex reasoning, whereas type~1 questions use a single SQL query. We observe an overall decline in average SER as task difficulty increases, decreasing from 79.47\% for type~1 to 59.16\% for type~2, 58.14\% for type~3, and 47.33\% for type~4. This monotonic downward trend culminates in a pronounced drop for type~4 questions, highlighting the greater challenge of generating multiple executable SQL queries for higher-level reasoning tasks. By our evaluation framework, we observe no similar trend.

\paragraph{Finding 3: Open-source LLMs particularly struggle with type~2 to type~4 questions.}
Table~\ref{tab:big-eval-3evals} shows that open-source LLMs consistently struggle with type~2--4 questions. Llama~4 and Qwen~3 achieve substantially lower performance than closed-source LLMs across type~2--4 questions. In particular, the average scores for open-source models reach 1.82 on type~2 questions, 2.45 on type~3 questions, and 2.35 on type~4 questions, indicating the substantially greater difficulty of generating SQL queries for open-ended BI tasks.

\section{Conclusion}
We introduce CORGI, a new text-to-SQL benchmark featuring 10 hand-curated industry-inspired databases, a new business literature-informed evaluation framework, and two new evaluation mechanisms. We show that popular LLMs struggle with increasingly difficult business insight questions such as those requiring recommendations. While LLMs are good at generating structured, MBA-trained responses, they fail to give sound implementable action plans. We also show that on average open-source models trail behind close-source models in SQL generation. We release CORGI as a public benchmark for researchers to contribute to and help to further our understanding of LLMs as management consultants.

\section*{Limitations}
While CORGI advances schema realism and task relevance for business applications, several limitations remain. 
First, all database contents and SQL queries are synthesized rather than drawn from production systems; this ensures privacy and scalability but omits natural artifacts such as noise, redundancy, and schema drift. Second, CORGI focuses on single-turn queries, leaving out multi-turn conversational refinement that is common in real-world systems. Third, each query is paired with only one natural language formulation, limiting the evaluation of model robustness to paraphrasing or informal inputs. Finally, this study covers only 10 business verticals. These limitations highlight promising directions for extending CORGI in future work.


\section*{Ethical Considerations}
Ethical considerations play an important role in our research. First, our evaluation framework incorporates an \textit{Ethical Responsibility} metric to assess whether generated analyses and recommendations adhere to established principles of AI ethics.  Second, we publicly release the dataset, evaluation questions, and evaluation framework on GitHub, enabling other researchers to replicate, scrutinize, and extend our work, thereby promoting transparency and collective progress in the text-to-SQL community. With respect to data privacy, all data used to construct the dataset are sourced exclusively from publicly available materials and do not contain confidential or sensitive information. Through these measures, we aim to introduce a new SQL evaluation benchmark while ensuring its responsible, transparent, and equitable use.

\section*{Acknowledgments}
We thank the anonymous reviewers and area chairs for their valuable feedback. We thank Tuan Huynh and Andrea W. Wen-Yi for their kind support.

\bibliography{custom}

\appendix
\section{Detailed Descriptions of CORGI Business Scenarios}
\label{sec:appendix-business}

We provide the following additional details for the 10 business scenarios covered in CORGI.

\paragraph{1. Online food delivery platform (e.g. Doordash)} This type of business connects users with a variety of local restaurants, and connects delivery personnel with different orders. The platform differentiates itself through personalized features including allergen tracking, prime club membership, customizable favorite lists, and family account management. Dynamic business features such as time-bound promotions and surge pricing allow restaurants to optimize engagement and profitability. Users can browse restaurants by categories, view detailed menus organized into sections and dishes, and place order. Orders progress through detailed status stages, and are fulfilled by a fleet of delivery personnel tracked by availability and vehicle type, with their performance recorded in a delivery history log. Users can rate not only restaurants but also individual dishes and delivery service, while restaurants can respond directly to reviews. The system supports secure payments, real-time tracking, promotional codes, and supports reviews and replies to ensure feedback loops.

\paragraph{2. App store (e.g. Google Play)} This type of business connects developers and users through app discovery, download, and monetization. Users can log in socially, maintain balances, and use multiple payment methods; developers can publish apps with version control, pricing models, and platform support; apps are categorized and tagged for enhanced discoverability; and usage is tracked through downloads, installations, rankings, and reviews.

\paragraph{3. Online clothing store (e.g. Lululemon)} This type of business usually operates an online and offline fashion retail platform that offers a diverse range of products categorized by department, category, and collection, and is tailored to various customer needs through uses, features, and fit types. The schema is centered on the Products table, which connects to departments, categories, collections, and is extended via junction tables for product uses and features. Each product links to one or more SKUs, which represent sellable items with inventory, pricing, and shipment details. Customer data is also extensively modeled, including multiple addresses, payment methods, loyalty membership, and order history, while promotional codes, returns, cancellations, and fulfillment options are designed to support scalable and efficient operations across both digital and physical retail channels.

\paragraph{4. Freelancing platform (e.g. Upwork)} This type of business operates a freelancing platform that connects employers and freelancers globally. Users can be differentiated between employers and freelancers. Employers can post jobs, each categorized and requiring specific skills. Freelancers can maintain a detailed profile with portfolio, education, experience and certifications. They can apply for the jobs that employers posted, which may lead to contracts. The wage is handled by different payment methods. After the completion of contracts, employers and freelancers can give reviews to each other.

\paragraph{5. Online room booking platform (e.g. Airbnb)} This type of business operates in the short-term rental and vacation housing market, connecting hosts who want to rent out properties with guests seeking accommodation. Users can act as either Hosts or Guests. Listings are tied to specific neighborhoods, cities, and countries through hierarchical foreign keys. Each listing includes details such as RoomTypes, pricing, amenities, availability, and policies. Booking activities are linked to users and listings, and supported by payment infrastructure (Payments, PaymentMethods, UserPaymentMethods). The system also tracks user engagement through Reviews and UserFavorites.

\paragraph{6. Personalized products (e.g. Persona Nutrition)} This type of business operates a platform that offers customized products tailored to individual customer needs. Users typically begin by completing a detailed assessment, which is then analyzed to provide personalized recommendations. For example, in the case of nutrition, the platform might offer custom vitamin packs, but this model can extend to other product categories such as skincare, beauty, and wellness products. Each product is linked to specific SKUs, representing items tailored to the customer's preferences. The business model is often subscription-based, ensuring regular delivery of the customized products. Additionally, the platform may offer reordering features, customer support for further customization, and a feedback loop to refine future product recommendations based on user reviews and changing needs.

\paragraph{7. Car Rental services (e.g.Turo)} This type of business operates a peer-to-peer rental platform that connects vehicle owners with individuals seeking short-term car rentals. Vehicle owners, known as hosts, can list their cars on the platform, specifying the car’s availability, location, and rental price. Renters, known as guests, can browse these listings, select a vehicle, and book it for the desired period. The platform charges a commission on each transaction, with the commission percentage varying based on factors like the type of vehicle and rental duration. Each rental is managed through the system, tracking reservations, pricing, insurance, and payment processing. Additionally, features like user verification, vehicle delivery, and 24/7 support ensure a seamless and secure experience for both parties. The platform also allows renters to rate hosts and vice versa after each transaction.

\paragraph{8. Second-hand luxury goods marketplace (e.g. TheRealReal)} This type of business operates a consignment platform specializing in the sale of pre-owned luxury goods, including clothing, handbags, jewelry, and art. Sellers consign their items to the platform, where experts authenticate each item before listing it for sale. The platform earns revenue by taking a commission from each sale, with rates varying based on the category and sale price of the item. The items are categorized and linked to detailed product pages, showcasing the brand, condition, and price. Buyers can browse and purchase authenticated luxury items from a wide range of high-end brands. The platform also offers additional services such as home pickup, personalized styling, and premium membership, enhancing the overall customer experience. Additionally, customers can sell their own pre-owned items through the platform, fostering a circular economy for luxury goods.

\paragraph{9. UX-driven advertising analytics platform (e.g. Mixpanel + Amplitude + AppsFlyer hybrid)} This type of business provides a full-stack data platform focused on optimizing digital advertising and customer experience through behavioral analytics, A/B testing, and multi-channel attribution. The platform integrates data from advertisement impressions, clicks, user sessions, content views, and CRM records to offer a unified view of the customer journey. Advertisers and product teams can launch campaigns, define audience segments, and test multiple ad creatives or landing pages using built-in experimentation tools. The platform supports real-time event tracking and visual funnel analysis, allowing businesses to identify drop-offs, optimize conversion paths, and personalize experiences at scale. The company also enables dynamic cohort building, personalized campaign delivery, and feedback-loop integration with external ad platforms (e.g., Meta, Google), helping brands make more informed, data-driven decisions to maximize ROI and user engagement.

\paragraph{10. E-commerce enablement platform (e.g. Shopify)} This type of business provides an end-to-end platform that enables entrepreneurs, SMBs, and large enterprises to launch, operate, and scale online storefronts with minimal technical complexity. The platform offers drag-and-drop website builders, integrated payment processing, inventory and order management, and multi-channel sales capabilities across marketplaces, social media, and in-person POS systems. It integrates data from product catalogs, customer interactions, and transaction histories to support a seamless retail experience. Merchants can run marketing campaigns, personalize product recommendations, and automate workflows using built-in analytics and third-party apps. The platform supports real-time inventory syncing, flexible fulfillment options, and automated tax and shipping calculations. Businesses can enhance customer engagement through CRM tools, loyalty programs, and abandoned cart recovery.

\section{Database Population Details}
\label{sec:data-generation-rules}

\paragraph{Step 1: Construct a schema.} 
We collect information from publicly available websites, applications, and technical documents to extract domain-specific workflows and business logic across diverse verticals. Based on this information, we construct entity–relationship diagrams and schemas that approximate the operational structures of real-world business.

\paragraph{Step 2: Simulate data}  
To capture complexities beyond static schemas and to approximate realistic commercial scenarios, we introduce three types of simulation rules to guide data generation. These rules are modeled in isolation as a simplifying assumption; no compounding effects are considered. This serves as the baseline for our study.\footnote{Future work will model interactions across rules.} For example, payment method distributions are not linked to seasonal trends: during Black Friday simulations, the proportions of payment methods are generated randomly rather than conditioned on sales surge. To illustrate the rules \textbf{a,b,c}, we simulate \textit{Persona Nutrition}, a personalized health and wellness subscription service, as a case study.  

    \textbf{(a) Business Operation Rules.} These are deterministic rules that directly reflect the company's operational logic. At \textit{Persona Nutrition}, new customers must complete (i) an \textit{Initial Assessment}. If the health assessment score falls below 3.0, (ii) a \textit{Low Rating Trigger} automatically generates consultation offers. Finally, (iii) \textit{Follow-up Assessments} are conducted every six months or upon user request. These operational constraints are kept consistent with the underlying schemas. 

    \textbf{(b) Latent Feature Distribution.} Demographic and behavioral patterns are instantiated through predefined distributions. For example, figure~\ref{fig:latent-feature} illustrates that at \textit{Persona Nutrition}, the customer base consists of 35\% trial users (0–3 months, price-sensitive), 44\% regular subscribers (3–12 months, consistent users), 17\% loyal customers (1–3 years, high engagement), and 4\% VIP customers (3+ years, premium services).

    \textbf{(c) Latent Seasonal Trends.} Our data also mimic realistic fluctuations. Figure~\ref{fig:latent-trend} illustrates these seasonal dynamics at \textit{Persona Nutrition}: subscriptions increase during events such as New Year's resolutions (+45\%) and Black Friday promotions (+27\%), while decreases are observed during periods such as post-New Year resolution (–33\%) and summer vacation (–31\%).

We employ Gemini-2.5-flash-lite\footnote{Version date: Sept 2025} \cite{comanici2025gemini25pushingfrontier} to synthesize data consistent with defined schemas and simulation rules. To ensure fidelity, all generated data is manually reviewed and randomly tested. Appendix \ref{sec:appendix-prompts} provides an example prompt used for data generation.


\section{Prompts}
\label{sec:appendix-prompts}
\subsection{Data Simulation Rules Generation}
The following is the prompt for generating data simulation rules, using Shopify as an example.

\begin{tcolorbox}[enhanced,
    breakable,
    colback=gray!3,        
    colframe=black!40,     
    colbacktitle=gray!15,  
    coltitle=black,        
    fonttitle=\bfseries,   
    title= Prompt for Generating Data Simulation Rules] 

\textbf{INSTRUCTIONS:} Assume you are a data scientist working on a text-to-SQL benchmark project. You now need to generate data simulation rules for Shopify database. To better approximate real business scenario of Shopify, you have: (1) Shopify description; (2) Link to Shopify official website; (3) Schema (\textbf{\textit{shopify.sql}}); (4) Real-world user experience report. You can retrieve information from reliable online sources.

\vspace{1em}
\textbf{OBJECTS:} Your need to design simulation rules three categories of data simulation rules: (1) Business Operation Rules: Deterministic rules that directly encode Shopify’s operational logic; (2) Latent Feature Distributions: Probabilistic rules capturing demographic and behavioral regularities; (3) Latent Trends: Temporal dynamics that reflect realistic fluctuations (e.g., seasonal demand shifts). 

\vspace{1em}
\textbf{REQUIREMENTS:} All rules should be documented in \textbf{\textit{rules.md}}, clearly delineated by category. When constructing the rules, adhere strictly to the tables, columns, and relationships explicitly defined in \textbf{\textit{shopify.sql}}. Do not introduce assumptions about missing or inferred schema elements.
\end{tcolorbox}

\subsection{Database Population}
The following is the prompt for populating database, using Shopify as an example.

\begin{tcolorbox}[enhanced,
    breakable,
    colback=gray!3,        
    colframe=black!40,     
    colbacktitle=gray!15,  
    coltitle=black,        
    fonttitle=\bfseries,   
    title= Prompt for Populating Database]

\textbf{INSTRUCTIONS:} Assume you are a data scientist working on a text-to-SQL benchmark project. You now need to populate the Shopify database. You are provded with: ((1) Schema (\textbf{\textit{shopify.sql}}); (2) Data simulation rules (\textbf{\textit{rules.md}}).

\vspace{1em}
\textbf{OBJECTS:} You should populate the Shopify database by strictly following \textbf{\textit{shopify.sql}} and \textbf{\textit{rules.md}}. You should populate the database as realistic as possible, simulating the real e-commerce business enablement platform scenario. 

\vspace{1em}
\textbf{REQUIREMENTS:} The database should be stored in \textbf{\textit{shopify.db}}.
\end{tcolorbox}

\subsection{Answer Generation}
The following are the prompts for answer generation.
\begin{tcolorbox}[enhanced,
    breakable,
    colback=gray!3,        
    colframe=black!40,     
    colbacktitle=gray!15,  
    coltitle=black,        
    fonttitle=\bfseries,    
    title= {Prompt for SQL Queries Generation}]

\textbf{INSTRUCTIONS:} You are an expert SQL analyst. Given a database schema and a business question, generate multiple comprehensive SQL queries (up to \{max\_queries\}) that will help answer the question,

\vspace{1em}
\textbf{GUIDELINES:} (1) Generate at most \{max\_queries\} different SQL queries that explore the question comprehensively; (2) Return each query in a separate code block with a brief description.

\vspace{1em}
Format your response as:

Query 1: [Brief description]

\begin{verbatim}
SELECT ...
FROM ...
WHERE ...
\end{verbatim}

\vspace{0.5em}
Query 2: [Brief description]

\begin{verbatim}
SELECT ...
FROM ...
WHERE ...
\end{verbatim}

And so on.

\vspace{1EM}
\textbf{User Prompt:}

Database Schema:
\{schema\}

Question: \{question\}

Please generate multiple SQL queries to comprehensively answer this question:
\end{tcolorbox}

\begin{tcolorbox}[enhanced,
    breakable,
    colback=gray!3,        
    colframe=black!40,     
    colbacktitle=gray!15,  
    coltitle=black,        
    fonttitle=\bfseries,    
    title= {Prompt for Final Answer Generation}]

\textbf{INSTRUCTIONS:} You are a senior data analyst. Given a business question and multiple SQL queries with their results, provide a comprehensive analysis that synthesizes all the information.

\vspace{1em}
Original Question: \{question\}

All Queries and Results:
\{query\_list\}, \{result\_list\}
\end{tcolorbox}

\subsection{Evaluation}
The following is the prompt for metric evaluation.
\begin{tcolorbox}[enhanced,
    breakable,
    colback=gray!3,        
    colframe=black!40,     
    colbacktitle=gray!15,  
    coltitle=black,        
    fonttitle=\bfseries,  
    title= {Prompt for Metric Evaluation},]

\textbf{INSTRUCTIONS:} You are an expert evaluator. Please evaluate the following answer based on the given question using the \{metric\} metric.

\vspace{1em}
Question: \{question\}
Answer: \{answer\}
Metric Definition: \{metric\}: \{metric definition\}
Evaluation Criteria: \{criteria\}

\vspace{1em}
\textbf{REQUIREMENTS:} Please provide your evaluation in the following JSON format:

\{\{

    "Score": <score 0-5>,
    
    "Reasoning": "<brief explanation of the score>"
    
\}\}

Please be precise and provide clear reasoning for your score. The output format should be strictly as above. Do not include any other text other than the JSON format. 
\end{tcolorbox}

\section{Complete Strategic Management Consulting-based Evaluation Framework}
\label{consulting-rubic}
Tables~\ref{tab:rubic1} and~\ref{tab:rubic2} present the complete version of strategic management consulting--inspired evaluation framework.

\newcolumntype{P}[1]{>{\raggedright\arraybackslash}p{#1}}

\clearpage
\pagestyle{empty}
\thispagestyle{empty}

\begin{table*}[t]
\centering
\footnotesize
\caption{\makebox[\linewidth]{Detailed Evaluation Rubric (Part 1)}}
\label{tab:rubic1}
\setlength{\tabcolsep}{4pt}
\renewcommand{\arraystretch}{1.25}
\emergencystretch=1em
\resizebox{\linewidth}{!}{%
\begin{tabular}{
P{2.0cm}
P{2.0cm}
P{2.0cm}
P{3.0cm}
P{3.7cm}
P{3.7cm}
P{3.7cm}
P{2.8cm}
}
\toprule
\textbf{Category} &
\textbf{Metric} &
\textbf{Scope} &
\textbf{Definition} &
\multicolumn{3}{c}{\textbf{Scores and Criteria}} &
\textbf{Sources} \\
\cmidrule(lr){5-7}
& & & &
\textbf{0 (Fails)} &
\textbf{3 (Meets)} &
\textbf{5 (Exceeds)} &
\\
\midrule

\multirow{3}{*}{\textbf{Structure}}
& Argument Soundness
& General
& Whether the answer's subpoints are Mutually Exclusive and Collectively Exhaustive (MECE), a core part of the Minto Pyramid Principle.
& The response is disorganized, lacks a main idea, and has redundant or illogical subpoints, making it not MECE. \textit{Example}: ``Our sales are down. We need to focus on marketing and then also on our customer base, especially the new ones. Also, products.''
& The response has a central idea but is only partially MECE; subpoints may overlap or not be fully exhaustive. The logic requires some effort to follow. \textit{Example}: ``Sales are down because of marketing and sales team performance. We should also look at our promotional strategies.''
& The response is a perfectly coherent pyramid, with a single main point and subpoints that are perfectly MECE and easy to grasp. \textit{Example}: ``Sales are down due to Price and Volume. We will analyze each of these areas separately to find the root cause.''
& Minto, B. (2009). The Minto Pyramid Principle: Logic in Writing, Thinking, and Problem Solving.\\

& Logical Coherence
& General
& (1) Contextual alignment: To assess whether the answer aligns with the context of the question (e.g., timeframe, entities). (2) Reasoning alignment: To assess whether the chain of reasoning is complete and sound.
& The response contradicts the context or is internally contradictory, with fundamentally flawed or incomplete reasoning. \textit{Example}: ``Our Q3 sales were 100M dollar, which is a 10\% decrease from last quarter's 80M dollar.''
& The response is mostly consistent but has minor logical gaps, making it difficult to follow. \textit{Example}: ``Our Q3 sales grew significantly because we launched a new product, which increased our market share.''
& The response is perfectly aligned with the context and has a complete, sound, and seamless chain of reasoning. \textit{Example}: ``Our Q3 sales grew by 10\% from Q2, from 80M to 88M dollar, which was a direct result of our new marketing campaign targeting urban areas.''
& Cosentino, M.P., 2024. Case in Point 12: Complete Case Interview. Burgee Press. \\

& Verbosity
& General
& Whether the answer is concise (minimum words, maximum information).
& The response is overly verbose, repetitive, and uninformative. \textit{Example}: ``In order to improve our sales performance, it is imperative that we strategically and proactively implement a new initiative designed to enhance overall revenue generation and expand our presence in the market.''
& The response is of reasonable length but could be more concise, containing some redundant details. \textit{Example}: ``To improve our sales, we should put a new initiative in place to increase revenue and market presence.''
& The response is concise and to the point, using the minimum words to convey maximum information efficiently. \textit{Example}: ``To increase sales, we will launch a new initiative to expand our market presence.''
& Garner, B.A., 2013. HBR guide to better business writing (HBR Guide Series). Harvard Business Review Press. \\
\addlinespace[6pt]
\bottomrule

\multirow{3}{*}{\textbf{Data Sense}}
& Information Relevance and Adequacy
& General
& Whether internal/external information is relevant and sufficient to answer the question and user intent.
& The response is irrelevant or inadequate, missing critical information or misinterpreting the user's intent. \textit{Example}: ``The Q3 revenue decline was caused by the cancellation of an insignificant user's account.''
& The response is functional but misses critical details or context, leaving significant follow-up questions. \textit{Example}: ``The Q3 revenue decline was caused by decreased sales in the US market.''
& The response is relevant and comprehensive, providing all necessary information to fully address the query and anticipate user needs. \textit{Example}: ``The Q3 revenue decline was due to a 10\% drop in unit volume, driven by increased competition and a new competitor's product launch that targeted our key demographic.''
& Voorhees, E. M. (2001). The TREC question answering track. Natural Language Engineering, 7(4), 361-378.\\

& Trend Awareness
& General
& Whether relevant trends (internal/external) are identified and integrated into the analysis.
& The response is static, providing data without considering trends. \textit{Example}: ``The sales in next November will be the same as October.'' (Igoring Black Friday trend)
& The response acknowledges trends but fails to integrate them meaningfully into the analysis. \textit{Example}: ``Next month is holiday season, but we do not need to take this into account when managing inventories.''
& The response acknowledges trends but fails to integrate them meaningfully into the analysis. \textit{Example}: ``Next month is holiday season, but we do not need to take this into account when managing inventories.''
& Binder, J. and Watkins, M.D., 2024. To solve a tough problem, reframe it. Harvard Business Review, 102(1-2), pp.80-89. \\

& Model Selection Rationality
& Unique -- Prediction (numerical)
& Whether a rational prediction model is selected, justified, and limitations acknowledged.
& An inappropriate or unsuited model is used for the prediction, showing a lack of fundamental data science understanding. \textit{Example}: ``We used a simple linear regression model to predict next year's sales, ignoring seasonality and market shifts.''
& A reasonable model is selected, but the choice is not justified, and its limitations are not acknowledged. \textit{Example}: ``We used an ARIMA model to predict sales.''
& A rational and well-justified model is selected for the prediction, with a clear explanation of its reasoning and limitations. \textit{Example}: ``We chose a Prophet model to forecast sales due to its strong performance on time-series data with clear seasonal and holiday effects, which are present in our dataset.''
& Brea, C., Anderson, J., Bartling, R., Mayo, R., \& Mueller, F. (2018). Demand forecasting with advanced analytics. Bain \& Company.\\ 
\addlinespace[6pt]
\bottomrule

\multirow{3}{*}{\textbf{Insightfulness}}
& Out-of-Box Thinking
& General
& Whether the analysis introduces novel concepts/heuristics beyond standard frameworks.
& The reasoning is conventional, relying only on standard formulas or predetermined steps. \textit{Example}: ``Used the standard 5 Forces analysis without adjustment.''
& The reasoning adapts a standard framework to the problem context but does not introduce a novel concept. \textit{Example}: ``Used a modified Profitability Tree to incorporate a new variable.''
& The response develops a novel analytical approach or heuristic to solve the problem, going beyond expected logical steps. \textit{Example}: ``Developed a new heuristic to model customer retention based on unstructured social data, resulting in a novel segmentation.''
& Backler, W., Iny, A., D’Intino, N., Parker, E., \& Hirashita, S. (2025, January 14). Navigating the future with strategic foresight. Boston Consulting Group. \\

& Root Cause Depth
& Unique -- Explanation
& Whether the analysis identifies plausible, well-supported root causes (beyond symptoms).
& The response only describes symptoms and offers no root cause analysis. \textit{Example}: ``Sales are down because customers are not buying our products.''
& The analysis is shallow or based on unsupported assumptions, providing weak conclusions. \textit{Example}: ``Sales are down due to a lack of marketing and high prices.''
& The analysis is thorough and multi-faceted, identifying plausible, well-supported root causes and proposing mitigations. \textit{Example}: ``Sales are down due to a new competitor that entered the market with lower prices and a superior product. This has made our existing product non-competitive, leading to a 20\% drop in volume.''
& inder, J. and Watkins, M.D., 2024. To solve a tough problem, reframe it. Harvard Business Review, 102(1-2), pp.80-89.\\

& Assumption Appropriateness
& Unique -- Prediction
& Whether assumptions are realistic, stated/justified, and conclusions match assumptions.
& The analysis is based on unrealistic or unstated assumptions, invalidating the conclusion. \textit{Example}: ``Our new product will be a success.'' (Based on no data)
& Some assumptions are appropriate but are not clearly stated or justified, leading to a confusing analysis. \textit{Example}: ``We assume that Gen Z will respond positively to our new product due to its viral potential.''
& All assumptions are clearly stated, justified, and realistic, with conclusions that logically follow and are fully consistent. \textit{Example}: ``We assume a 5\% increase in market share in Q4, based on our successful pilot program and customer feedback showing a high intent to purchase.''
& Mulligan, C., Northcote, N., Röder, T., \& Vesuvala, S. (2021, September 9). The strategy-analytics revolution. McKinsey \& Company.  \\
\addlinespace[6pt]
\bottomrule
\end{tabular}%
}
\end{table*}

\clearpage
\pagestyle{empty}
\thispagestyle{empty}

\begin{table*}[t]
\centering
\footnotesize
\caption{\makebox[\linewidth]{Detailed Evaluation Rubric (Part 2)}}
\label{tab:rubic2}
\setlength{\tabcolsep}{4pt}
\renewcommand{\arraystretch}{1.25}
\emergencystretch=1em
\resizebox{\linewidth}{!}{%
\begin{tabular}{
P{2.0cm}
P{2.0cm}
P{2.0cm}
P{3.0cm}
P{3.7cm}
P{3.7cm}
P{3.7cm}
P{2.8cm}
}
\toprule
\textbf{Category} &
\textbf{Metric} &
\textbf{Scope} &
\textbf{Definition} &
\multicolumn{3}{c}{\textbf{Scores and Criteria}} &
\textbf{Sources} \\
\cmidrule(lr){5-7}
& & & &
\textbf{0 (Fails)} &
\textbf{3 (Meets)} &
\textbf{5 (Exceeds)} &
\\
\midrule
\multirow{2}{=}{\textbf{Operation Implementability}}
& Actionability
& Unique -- Recommendation
& Whether recommendations are executable, concrete, and tied to measurable steps or KPIs.
& The recommendation is too vague or abstract to be executed, with no concrete steps. \textit{Example}: ``The company should improve its online presence.''
& The recommendation is high-level and lacks specific, measurable steps for implementation. \textit{Example}: ``The company should launch a social media campaign and update its website.''
& The recommendation is highly actionable, concrete, and includes specific steps tied to business KPIs. \textit{Example}: ``Launch a TikTok campaign with five influencer partners, with a budget of 50,000 dollar, and redesign the homepage with A/B testing on two new layouts.''
& Garner, B.A., 2013. HBR guide to better business writing (HBR Guide Series). Harvard Business Review Press. \\

& Time-Based Planning
& Unique -- Recommendation
& Whether recommendations include a realistic implementation timeline with clear milestones.
& The recommendation lacks any timeline or sense of urgency. \textit{Example}: ``We recommend a new marketing campaign.''
& The recommendation includes a vague, general timeline that does not allow for effective planning. \textit{Example}: ``We will launch the new campaign in the near future.''
& The recommendation has a specific, realistic, and well-justified timeline with clear milestones for effective planning. \textit{Example}: ``We will launch the new marketing campaign in Q4, with a goal of completion by December 15th.''
& Brea, C., Anderson, J., Bartling, R., Mayo, R., \& Mueller, F. (2018). Demand forecasting with advanced analytics. Bain \& Company.  \\
\addlinespace[6pt]
\midrule
\multirow{2}{=}{\textbf{Purpose Alignment}}
& Goal Orientation
& Unique -- Recommendation
& Whether the recommendation is linked to reachable, measurable goals (SMART) and a clear value proposition.
& The recommendation is not linked to any clear goals or business value. \textit{Example}: ``We suggest you hire a new VP of Marketing.''
& The recommendation is linked to a general goal that is not specific or measurable. \textit{Example}: ``We suggest you hire a new VP of Marketing to increase brand awareness.''
& The recommendation is directly tied to a specific, measurable, and achievable (SMART) business goal with a clear value proposition. \textit{Example}: ``We suggest you hire a new VP of Marketing with the goal of increasing our market share by 5\% and our social media engagement by 20\% in the next fiscal year.''
& Harvard Business Review Analytic Services. (2022). Implementing strategic goals for organizational success. Harvard Business School Publishing. Sponsored by Lattice \\

& Stakeholder Orientation
& Unique -- Recommendation
& Whether the recommendation considers and balances impacts on key non-shareholder stakeholders.
& The recommendation focuses exclusively on financial metrics (e.g., ROI) and ignores all non-shareholder impacts. \textit{Example}: ``Recommend mass layoffs to cut costs by 20\% without considering employee morale or retention.''
& Mentions key stakeholders (e.g., customers) but does not integrate their perspectives into the final recommendation or mitigation plan. \textit{Example}: ``The plan may impact employees, but the financial return justifies it.''
& The recommendation explicitly balances financial gains with stakeholder well-being, including measures to mitigate negative effects on non-shareholder groups. \textit{Example}: ``Recommend a phased automation rollout, offsetting job losses with a robust retraining program for high-value roles.''
& Bain \& Company. (2023, January). Stakeholder strategies – Management tools. ) \\
\addlinespace[6pt]
\midrule
\multirow{3}{*}{\textbf{Compliance}}
& Risk Management
& Unique -- Recommendation
& Whether risks are identified, warned, and accompanied by concrete mitigation strategies.
& The response fails to identify any risks or contains content that exacerbates risk. \textit{Example}: ``The new product launch will be a guaranteed success.''
& Some risks are identified, but warnings or mitigation suggestions are inadequate. \textit{Example}: ``A new product launch could fail, so we should be careful.''
& The response provides comprehensive risk warnings, identifies potential risk points, and offers concrete mitigation suggestions. \textit{Example}: ``We recognize the risk of market saturation and mitigate this by conducting a thorough competitive analysis, allocating a contingency budget for unexpected costs, and a phased rollout to a test market.''
& Guo, J., Guo, Y., Li, M., \& Tan, S. (2025). Flame: Financial large-language model assessment and metrics evaluation. arXiv preprint arXiv:2501.06211. \\

& Regulatory Compliance
& Unique -- Recommendation
& Whether recommendations adhere to applicable laws, regulations, and industry standards, including disclaimers.
& The response is non-compliant with relevant laws and regulations, posing a severe risk. \textit{Example}: ``To bypass data privacy regulations, you should use customer data from our partners without their explicit consent.''
& The response is generally compliant but lacks specific disclaimers or fails to fully address all regulatory requirements. \textit{Example}: ``Our new data collection process will adhere to all regulations, but we will not be adding a legal disclaimer to our website.''
& The response is fully compliant with all applicable laws and regulations, including adequate disclaimers and privacy protocols. \textit{Example}: ``All of our new data collection processes are fully compliant with GDPR and CCPA, and we have implemented a clear, user-facing privacy policy and consent form to ensure full legal adherence.''
& Guo, J., Guo, Y., Li, M., \& Tan, S. (2025). Flame: Financial large-language model assessment and metrics evaluation. arXiv preprint arXiv:2501.06211. \\

& Ethical Responsibility
& Unique -- Recommendation
& Whether recommendations uphold AI ethics (e.g., fairness, non-discrimination, well-being) with concrete actions.
& The recommendation is demonstrably biased or perpetuates known social inequalities, or ignores explicit ethical consequences. \textit{Example}: ``The loan risk model shows bias against applicants from specific zip codes; recommend proceeding without mitigation.''
& Includes a general statement about ``fairness'' but lacks concrete, measurable steps to mitigate identified bias or ensure equitable outcomes. \textit{Example}: ``We should try to be fair, but the model output is final.''
& The recommendation includes specific, actionable steps (e.g., bias monitoring, disparate impact analysis) to ensure equitable outcomes and promote the well-being of the affected demographic. \textit{Example}: ``Implement a specific re-weighting algorithm to address observed demographic bias in the output and monitor fairness metrics in real time.''
& Cowgill, B., Dell'Acqua, F., Deng, S., Hsu, D., Verma, N. and Chaintreau, A., 2020, July. Biased programmers? Or biased data? A field experiment in operationalizing AI ethics. In Proceedings of the 21st ACM Conference on Economics and Computation (pp. 679-681). \\
\addlinespace[6pt]
\midrule
\end{tabular}%
}
\end{table*}
\clearpage

\section{Evaluation Examples}
\label{sec:appendix-case}

\begin{tcolorbox}[enhanced,
    breakable,
    colback=gray!3,        
    colframe=black!40,     
    colbacktitle=gray!15,  
    coltitle=black,        
    fonttitle=\bfseries,   
    title= {Type 2: Why did download numbers in March 2025 much higher than other months? (Appstore db)},]
\textbf{Responder: Gemini-2.5-flash-lite}

The data reveals a significant spike in app downloads in March 2025, with \textbf{3006 downloads} compared to an average of approximately \textbf{1800--1900 downloads} in surrounding months (Query 1). This analysis investigates the contributing factors behind this surge.

\paragraph{Key Drivers of the Download Increase:}

\begin{itemize}
    \item \textbf{Dominance of ``BlackMonkey'' App:} The primary driver of the download surge in March 2025 was overwhelmingly the app \textbf{``BlackMonkey''}, which accounted for \textbf{1222 downloads} (Query 2). This single app represents a substantial portion of the total downloads for that month, far exceeding any other individual app.
    
    \item \textbf{Developer ``KidsCloud'' Success:} The developer \textbf{``KidsCloud''} appears to be closely linked to this surge, reporting \textbf{1208 downloads} in March 2025 (Query 7). Given that ``BlackMonkey'' is the top-downloaded app and ``KidsCloud'' is the top-downloaded developer, it is highly probable that ``BlackMonkey'' is developed by ``KidsCloud'' and is the reason for their exceptional download numbers.
    
    \item \textbf{Category Performance:} The surge is concentrated within the \textbf{``Sports''} (1364 downloads) and \textbf{``Food \& Drink''} (1251 downloads) categories (Query 4). This suggests that ``BlackMonkey'' (or any other significant contributors from KidsCloud) likely falls into one or both of these categories.
    
    \item \textbf{Age Group Engagement:} The ``Teen'' (\textbf{1824 downloads}) and ``Kids'' (\textbf{1820 downloads}) age groups were the most active in March 2025 (Query 5). This aligns with the hypothesis that ``BlackMonkey'' is an app popular with younger demographics, potentially a game or an entertainment app.
    
    \item \textbf{Platform Distribution:} While iOS and Android show high download numbers (\textbf{2307} and \textbf{2299} respectively), macOS and Windows also saw a considerable increase in downloads compared to other months (Query 3). This indicates that the surge in downloads for ``BlackMonkey'' (and potentially other apps from KidsCloud) was not limited to a single platform.
    
    \item \textbf{Pricing Model:} Both \textbf{``paid''} (\textbf{1284 downloads}) and \textbf{``free''} (\textbf{1200 downloads}) apps saw a significant increase in downloads, with subscription models also showing a moderate rise (\textbf{522 downloads}) (Query 6). This suggests that the increased downloads might be a mix of paid app popularity and free app adoption, or that ``BlackMonkey'' itself might have a tiered pricing model or in-app purchases.
\end{itemize}

\paragraph{App Releases and Updates in March 2025:} Query 8 lists several app releases and updates in March 2025. While many popular apps like PicsArt, Nike Training Club, and Clash of Clans had updates, none of these individually appear to have the download volume to explain the overall surge. The app \textbf{``BlackMonkey''} does not appear in the list of updated or released apps in March 2025. This suggests that either ``BlackMonkey'' had no significant update or release in March, or that the surge is driven by an app not captured in this specific query (e.g., an app with a very long name not easily identifiable, or an app released prior to March but gaining significant traction due to external factors).

\paragraph{Conclusion:} The significant increase in app downloads in March 2025 was primarily driven by the app \textbf{``BlackMonkey''}, likely developed by \textbf{``KidsCloud''}. This app appears to be particularly popular among \textbf{teen and kid} demographics and has likely contributed heavily to the \textbf{Sports} and \textbf{Food \& Drink} categories. The download surge was observed across major platforms (iOS, Android, macOS, Windows) and was influenced by both paid and free app downloads. Further investigation could focus on understanding the specific marketing efforts or viral nature of ``BlackMonkey'' that led to this exceptional performance.

\noindent \textbf{Evaluator: GPT-4o}

\noindent Scores:
\begin{itemize}
    \item Structure
    \begin{itemize}
        \item Argument Soundness: 5
        \item Logical Coherence: 5
        \item Verbosity: 3
    \end{itemize}
    \item Data Sense
    \begin{itemize}
        \item Information Adequacy: 5
        \item Trend Awareness: 3
    \end{itemize}

    \item Insightfulness
    \begin{itemize}
        \item Out-of-box Thinking: 3
        \item Root Cause Depth: 5
    \end{itemize}
    
\end{itemize}
\end{tcolorbox}

\begin{tcolorbox}[enhanced,
    breakable,
    colback=gray!3,        
    colframe=black!40,     
    colbacktitle=gray!15,  
    coltitle=black,        
    fonttitle=\bfseries,   
    title= {Type 3: What will our return rate be during the next holiday season? (Clothing E-commerce DB)},]
\textbf{Responder: GPT-4o}

\paragraph{Comprehensive Analysis.} The original question seeks to estimate the return rate for the next holiday season based on historical data. By synthesizing the results of the provided queries, we can derive insights into the overall return rate, the primary reasons for returns, the categories with the highest return rates, and the impact of customer loyalty tiers on returns. These insights will help us predict and potentially mitigate returns during the next holiday season.

\subsubsection*{1. Overall Return Rate (Query 1)}
The overall return rate during the last holiday season (November 1, 2022, to December 31, 2022) was \textbf{11.52\%}. Out of 382 total orders, 44 were returned. This provides a baseline for estimating the return rate for the upcoming holiday season. If similar conditions persist (e.g., product offerings, customer base, and operational factors), we can expect a return rate in the range of \textbf{11--12\%}.

\subsubsection*{2. Primary Reasons for Returns (Query 2)}
The analysis of return reasons reveals the following key insights:
\begin{itemize}
    \item The top reasons for returns were \textbf{``Didn't fit'' (26.47\%)}, \textbf{``Wrong size'' (23.53\%)}, and \textbf{``Quality issue'' (23.53\%)}. Together, these three reasons accounted for over 73\% of all returns.
    \item Other reasons, such as \textbf{``Defective item'' (14.71\%)} and \textbf{``Changed mind'' (11.76\%)}, were less frequent but still notable.
\end{itemize}

This suggests that sizing and quality are the primary drivers of returns. To reduce the return rate, the company could:
\begin{itemize}
    \item Improve size guides and provide more detailed product descriptions.
    \item Enhance quality control processes to minimize defective items.
    \item Offer virtual fitting tools or customer reviews to help customers make more informed decisions.
\end{itemize}

\subsubsection*{3. Return Rates by Product Category (Query 3)}
The return rates varied significantly across product categories:
\begin{itemize}
    \item \textbf{Dresses (16.96\%)}, \textbf{Swimwear (15.70\%)}, and \textbf{Shoes (13.97\%)} had the highest return rates. These categories are likely more prone to sizing and fit issues, which aligns with the reasons for returns identified in Query 2.
    \item Categories like \textbf{Bottoms (8.45\%)} and \textbf{Formal Wear (10.00\%)} had lower return rates.
\end{itemize}

This indicates that the company should focus on addressing fit-related concerns in high-return categories like Dresses, Swimwear, and Shoes. For example:
\begin{itemize}
    \item Offering free exchanges for different sizes in these categories could reduce the likelihood of outright returns.
    \item Providing detailed size charts and customer reviews specific to these categories could also help.
\end{itemize}

\subsubsection*{4. Impact of Customer Loyalty Tiers on Returns (Query 4)}
The return rates by loyalty tier were as follows:
\begin{itemize}
    \item \textbf{Silver (14.47\%)} and \textbf{Gold (14.12\%)} customers had the highest return rates.
    \item \textbf{Platinum (7.50\%)} customers had the lowest return rate, followed by \textbf{Bronze (10.39\%)} and customers with no loyalty tier (10.94\%).
\end{itemize}

This suggests that higher-tier loyalty customers (Silver and Gold) are more likely to return items, possibly because they purchase more frequently or feel more confident in the return process. Platinum customers, on the other hand, may be more selective in their purchases, leading to fewer returns.

To address this:
\begin{itemize}
    \item The company could analyze the purchasing behavior of Silver and Gold customers to identify patterns leading to higher returns.
    \item Offering personalized recommendations or targeted size/fit guidance for these tiers could help reduce returns.
\end{itemize}

\subsubsection*{5. Key Takeaways and Recommendations}
Based on the analysis, the following insights and actions can help predict and mitigate the return rate for the next holiday season.

\paragraph{Predicted Return Rate.}
The return rate for the next holiday season is likely to be in the range of \textbf{11--12\%}, consistent with the last holiday season.

\paragraph{Recommendations to Reduce Returns.}
\begin{enumerate}
    \item \textbf{Address Sizing and Fit Issues:}
    \begin{itemize}
        \item Focus on high-return categories like Dresses, Swimwear, and Shoes by improving size guides, offering virtual fitting tools, and providing detailed product descriptions.
        \item Consider offering free exchanges for size-related issues to reduce outright returns.
    \end{itemize}

    \item \textbf{Enhance Quality Control:}
    \begin{itemize}
        \item Since ``Quality issue'' accounted for 23.53\% of returns, improving quality checks and addressing common defects could significantly reduce returns.
    \end{itemize}

    \item \textbf{Targeted Interventions for Loyalty Tiers:}
    \begin{itemize}
        \item Analyze the purchasing behavior of Silver and Gold customers to identify patterns leading to higher returns.
        \item Provide personalized recommendations or targeted size/fit guidance for these tiers.
    \end{itemize}

    \item \textbf{Monitor and Address Seasonal Trends:}
    \begin{itemize}
        \item Since the holiday season often involves gifting, returns due to ``Changed mind'' or ``Wrong size'' may increase. Offering gift receipts or extended return windows could improve customer satisfaction while managing returns.
    \end{itemize}
\end{enumerate}

\noindent \textbf{Evaluator: Gemini-2.5-flash-lite}

\noindent Scores:
\begin{itemize}
    \item Structure
    \begin{itemize}
        \item Argument Soundness: 5
        \item Logical Coherence: 5
        \item Verbosity: 0
    \end{itemize}

    
    \item Data Sense
    \begin{itemize}
        \item Information Adequacy: 3
        \item Trend Awareness: 3
        \item Model Selection Rationale: 0
    \end{itemize}

    \item Insightfulness
    \begin{itemize}
        \item Out-of-box Thinking: 0
        \item Assumption Appropriateness: 3
        \item Root Cause Depth: 5
    \end{itemize}
    
\end{itemize}
\end{tcolorbox}

\begin{tcolorbox}[enhanced,
    breakable,
    colback=gray!3,        
    colframe=black!40,     
    colbacktitle=gray!15,  
    coltitle=black,        
    fonttitle=\bfseries,   
    title={Type 4: How can I increase the repeat purchase rate? (Shopify DB)},]
\textbf{Responder: GPT-4o}

\paragraph{Comprehensive Analysis.}
Based on the query results, there is no data for customers with repeat purchases (i.e., customers with more than one order). This indicates that all existing customers have made only a single purchase. The absence of repeat orders may result from several factors, including limited post-purchase engagement, lack of incentives to return, or misaligned marketing strategies. Below is an analysis of the findings and actionable recommendations to improve the repeat purchase rate.

\subsubsection*{1. Key Observations}

\textbf{(a) No Repeat Purchases.}  
Query 1 confirms that there are no customers with more than one order (\texttt{total\_orders = 1}). Subsequent queries (2–8) analyzing patterns such as discounts, campaigns, delivery, or refunds also return no results, since no repeat customers exist.

\textbf{(b) Marketing Campaigns.}  
Despite substantial ad spending (e.g., up to \$24{,}156.55), Query 5 shows that marketing campaigns have not generated repeat customers, suggesting a weak alignment between marketing efforts and retention goals.

\textbf{(c) Fulfillment and Delivery.}  
Query 6 reports no delivery delays for repeat customers; however, this finding is irrelevant given the absence of such customers.

\textbf{(d) Discounts and Promotions.}  
Query 4 reveals no data on discounts driving repeat purchases, implying that either discount-based campaigns are lacking or ineffective.

\textbf{(e) Product Offerings.}  
Query 3 indicates that no specific product variant or collection has encouraged repeat purchases.

\textbf{(f) Refunds.}  
Query 8 shows no refund data related to repeat customers—unsurprising given their absence—suggesting that refund issues are not the primary barrier to repeat purchases.

\subsubsection*{2. Recommendations to Increase Repeat Purchases}

\paragraph{1. Understand Customer Behavior (Root-Cause Analysis).}
\begin{itemize}
    \item Conduct post-purchase surveys or interviews to identify reasons customers do not return.
    \item Evaluate satisfaction with product quality, pricing, and service experience.
    \item Verify that product value propositions align with customer expectations.
\end{itemize}

\paragraph{2. Enhance Customer Experience.}
\begin{itemize}
    \item \textbf{Personalization:} Use purchase history and preferences to recommend relevant products and send follow-up emails.  
    \item \textbf{Onboarding:} Deliver a seamless first-purchase experience that builds trust and recall.  
    \item \textbf{Fulfillment:} Offer fast and reliable delivery to reinforce satisfaction.  
    \item \textbf{Quality Assurance:} Maintain consistent product quality to encourage repeat buying.  
\end{itemize}

\paragraph{3. Introduce Loyalty Programs.}
\begin{itemize}
    \item Implement a points-based system rewarding repeat purchases.  
    \item Provide tiered benefits (e.g., Bronze, Silver, Gold) to incentivize engagement.  
\end{itemize}

\paragraph{4. Run Retargeting and Re-engagement Campaigns.}
\begin{itemize}
    \item Re-engage one-time customers via email, SMS, or social media.  
    \item Highlight new arrivals, exclusive offers, and limited-time discounts.  
    \item Use abandoned-cart reminders to recover potential repeat buyers.  
\end{itemize}

\paragraph{5. Leverage Discounts and Incentives.}
\begin{itemize}
    \item Offer second-purchase discounts (e.g., “10\% off your next order”).  
    \item Provide free shipping or gifts for repeat customers.  
    \item Introduce bundles or subscriptions to foster habitual purchases.  
\end{itemize}

\paragraph{6. Improve Marketing Effectiveness.}
\begin{itemize}
    \item Shift focus from acquisition to retention-oriented marketing.  
    \item Target existing customers with retargeting ads and personalized messaging.  
    \item Evaluate the ROI of current channels (affiliate, social) and expand to email or partnership marketing.  
\end{itemize}

\paragraph{7. Expand Product Offerings.}
\begin{itemize}
    \item Introduce complementary or add-on products to encourage follow-up purchases.  
    \item Bundle products to increase perceived value and convenience.  
\end{itemize}

\paragraph{8. Build Trust and Brand Loyalty.}
\begin{itemize}
    \item Communicate brand values and share authentic customer stories.  
    \item Feature reviews or user-generated content to build credibility.  
    \item Engage actively with customers on social platforms.  
\end{itemize}

\paragraph{9. Monitor and Optimize.}
\begin{itemize}
    \item Track metrics such as repeat customer rate, average order value, and customer lifetime value.  
    \item Continuously experiment with retention tactics and optimize based on results.  
\end{itemize}

\subsubsection*{3. Immediate Next Steps}
\begin{enumerate}
    \item \textbf{Survey existing customers} to collect feedback on satisfaction and barriers to repeat purchases.  
    \item \textbf{Launch a “Welcome Back” campaign} offering targeted discounts to first-time buyers.  
    \item \textbf{Analyze marketing ROI} and reallocate budget toward retention activities.  
    \item \textbf{Design a loyalty program} to reward repeat behavior.  
\end{enumerate}

\noindent \textbf{Evaluator: GPT-4o}

\noindent \textbf{Scores:}
\begin{itemize}
    \item Structure
    \begin{itemize}
        \item Argument Soundness: 5
        \item Logical Coherence: 5
        \item Verbosity: 3
    \end{itemize}
    \item Data Sense
    \begin{itemize}
        \item Information Adequacy: 5
        \item Trend Awareness: 3
        \item Model Selection Rationale: 3
    \end{itemize}
    \item Insightfulness
    \begin{itemize}
        \item Out-of-box Thinking: 5
        \item Assumption Appropriateness: 3
    \end{itemize}
    \item Operation Implementability
    \begin{itemize}
        \item Actionability: 5
        \item Time-based Planning: 0
    \end{itemize}
    \item Purpose Alignment
    \begin{itemize}
        \item Goal Orientation: 5
        \item Stakeholder Orientation: 5
    \end{itemize}
    \item Compliance
    \begin{itemize}
        \item Risk Management: 5
        \item Regulatory Compliance: 3
        \item Ethical Responsibility: 5
    \end{itemize}
\end{itemize}
\end{tcolorbox}

\section{Implementation Details of Bradley--Terry Framework}
\label{bt-framework}
We use the Bradley--Terry framework to model pairwise preferences and implement it with the \texttt{choix} PyPI package (v0.3.6). 
We assign each item $i$ a latent strength parameter $\theta_i > 0$. 
For $n$ comparisons, we uniformly sample a pair $(i, j)$ from all $\binom{n}{2}$ possible pairs 
and determine the outcome through a Bernoulli trial with 
$P(i \succ j) = \frac{\theta_i}{\theta_i + \theta_j}$. 
We record $(a, b)$ when $a$ is preferred over $b$, 
producing a consistent dataset for ranking evaluation under stochastic preferences.



\section{Discussion of Aggregation Methods}
\label{aggregation}

We examine whether increasing the influence of the three dimensions specific to type~4 tasks, namely Operational Implementability (OI), Purpose Alignment (PA), and Compliance (C), alters the overall performance ranking for type~4 tasks. We consider two alternative weighting schemes: (i) \textit{Method A}, which doubles the weights of OI, PA, and C while keeping the weights of Structure, Data Sense, and Insightfulness unchanged; and (ii) \textit{Method B}, which assigns a weight of 30\% to each of OI, PA, and C, and 10\% to each of the remaining three dimensions.

Table~\ref{tab:summary-type4} shows that: (i) all models achieve substantially lower scores on type~4 tasks than on type~2 and type~3 tasks, and this trend persists under both weighting schemes A and B, with the performance drop being especially pronounced for GPT-4o~\cite{openai2024gpt4ocard} on type~4; (ii) the relative ranking of models remains stable across all task types (GPT-4o > Gemini-2.5-flash-lite > Gemini-2.0-flash-lite~\cite{google_gemini2.0}), and this ordering is fully consistent across type~2--4 tasks, including their weighted variants; and (iii) increasing the weights of the three type~4 dimensions does not alter the overall ranking.

\begin{table}[t]
\centering
\small
\setlength{\tabcolsep}{5pt}
\begin{tabularx}{\linewidth}{lccccc}
\toprule
\textbf{Model} & \textbf{T2} & \textbf{T3} & \textbf{T4} & \textbf{T4\_A} & \textbf{T4\_B} \\
\midrule
Gemini-2.0-flash-lite & 3.06 & 3.05 & 3.05 & 3.012 & 2.993 \\
Gemini-2.5-flash-lite & 3.08 & 3.10 & 3.08 & 3.042 & 3.026 \\
GPT-4o                & 3.38 & 3.32 & 3.27 & 3.244 & 3.231 \\
\bottomrule
\end{tabularx}
\caption{Impact of different aggregation methods. T denotes question type.}
\label{tab:summary-type4}
\end{table}


\section{Kendall's Tau ($\tau$)}
\label{appendix:kendall}

We use Kendall's tau ($\tau$) to assess the consistency between automatic evaluation scores and human expert judgments. Kendall's tau is a nonparametric statistic that measures the ordinal association between two ranked variables. It quantifies ranking consistency by comparing the relative ordering of all pairs of observations.

Given a set of paired observations $(x_i, y_i)$, Kendall's tau is defined as:
\begin{equation}
\tau = \frac{P - Q}{P + Q},
\end{equation}
where $P$ denotes the number of \emph{concordant} pairs and $Q$ denotes the number of \emph{discordant} pairs. 
A pair $(i, j)$ is concordant if the relative ordering of $x_i$ and $x_j$ agrees with that of $y_i$ and $y_j$, and discordant otherwise. 
The value of $\tau$ lies in the interval $[-1, 1]$, where $\tau = 1$ indicates perfect agreement, $\tau = -1$ indicates perfect disagreement, and $\tau = 0$ indicates no ordinal association.


\section{Additional Benchmark Statistics}
\label{appen-sta}
Table~\ref{tab:corgi-vs-benchmarks} presents additional statistics of the CORGI benchmark.

\begin{table}[t]
\centering
\small
\setlength{\tabcolsep}{4pt}
\renewcommand{\arraystretch}{1.1}
\caption{CORGI vs.\ other benchmarks. \#D and \#T/DB denote the number of databases and the average number of tables per database, respectively.}
\label{tab:corgi-vs-benchmarks}
\begin{tabular}{p{2.7cm}ccccc}
\toprule
\textbf{Dataset} & \textbf{\#D} & \textbf{\#T/DB} & \textbf{Domain} & \textbf{BI Question} \\
\midrule
Spider      & 138 & 5.1 & Cross  & $\times$ \\
WikiSQL     & --  & 1   & Cross  & $\times$ \\
BIRD        & 37  & 7.3 & Cross  & $\times$ \\
Advising    & 1   & 10  & Single & $\times$ \\
\midrule
\textbf{CORGI} & \textbf{10} & \textbf{26} & \textbf{Cross} & \checkmark \\
\bottomrule
\end{tabular}
\end{table}

\end{document}